\documentclass[journal,twoside,web]{ieeecolor}
\usepackage{generic}
\usepackage{cite}
\usepackage{amsmath,amsfonts,amssymb}
\usepackage{algorithm}
\usepackage{algorithmicx}
\usepackage{algpseudocode}
\usepackage{array}
\usepackage{textcomp}
\usepackage{url}
\usepackage{graphicx}

\usepackage{booktabs}
\usepackage{multirow}
\usepackage{makecell}
\usepackage{bm}
\usepackage[dvipsnames]{xcolor}
\usepackage[colorlinks=true,linkcolor=Cerulean,citecolor=Cerulean,urlcolor=Cerulean]{hyperref}
\usepackage{orcidlink}
\usepackage{cleveref}
\crefformat{figure}{Fig.~#2#1#3}
\crefformat{table}{Table~#2#1#3}
\crefformat{algorithm}{Algorithm~#2#1#3}
\crefformat{section}{#2Section~#1#3}
\crefformat{subsection}{#2Section~#1#3}
\usepackage{placeins}

\begin{document}

% 基于大语言模型引导的任务语义场分解：应用于工业过程预测
\title{LLM-Guided Task-Semantic Field Factorization for Industrial Process Forecasting}

\author{
  Youcheng Zong\(^{\orcidlink{0009-0008-6795-8412}}\),~\IEEEmembership{Student~Member,~IEEE},
  Runda Jia\(^{\orcidlink{0000-0002-8586-243X}}\),
  Mingxuan Ren\(^{\orcidlink{0009-0001-0172-5914}}\),
  \\ and Dakuo He\(^{\orcidlink{0000-0001-8303-529X}}\)
  \thanks{This work was supported by the Fundamental Research Funds for the Central Universities, China (N26GFZ006). \textit{(Corresponding author: Runda Jia.)}}
  \thanks{Youcheng Zong, Runda Jia, Mingxuan Ren, and Dakuo He are with the College of Information Science and Engineering, Northeastern University, Shenyang 110004, China (e-mail: youchengzong@stumail.neu.edu.cn; jiarunda@ise.neu.edu.cn; renmx@mails.neu.edu.cn; hedakuo@ise.neu.edu.cn).}
}

\maketitle

\begin{abstract}
  % 过程工业依赖时序预测和软测量来估计难以在线测量的质量变量。标注数据稀缺、工况变化频繁，为不同场景重新训练模型或重建对齐流程代价高昂。
  Process industries rely on time-series forecasting and soft sensing to estimate quality variables that are hard to measure online. Labeled data are scarce, operating regimes change frequently, and retraining models or rebuilding alignment pipelines for each scenario is costly.
  % 这些场景通常已有变量表和工艺文档，记录变量名称、单位、物理含义和工艺角色。然而，标准时序主干通常把输入看作匿名数值列。现有文本增强方法也很少将输入变量与预测目标之间的语义逻辑关系落实到每个数值窗口中的模型输入。
  Such settings often provide variable tables and process documents that record variable names, units, physical meanings, and process roles. However, standard time-series backbones usually treat inputs as anonymous numerical columns. Existing text-enhanced methods also rarely make the semantic-logical relations between input variables and the prediction target available to the model within each numerical window.
  % 为解决这一问题，本文提出任务语义场因子化（TSF），这是一种由大语言模型（LLM）引导的框架。TSF 在训练前从任务协议和变量文档中构建任务语义场，并且只在离线语义构造阶段使用 LLM。在线训练和推理由常规时序主干完成。训练和推理时，当前数值窗口激活变量语义，使语义信息参与每次预测，并支持模型面对不同预测目标和工况变化时进行适应。
  To address this problem, this article proposes Task-Semantic Field Factorization (TSF), a large language model (LLM)-guided framework. TSF builds a task-semantic field from task protocols and variable documents before training and uses the LLM only for offline semantic construction. Online training and inference are handled by conventional time-series backbones. During training and inference, the current numerical window activates variable semantics, so semantic information participates in each prediction and supports adaptation to different prediction targets and operating shifts.
  % 在多个复杂工业预测和延迟软测量任务上，TSF 的平均 MAE 降幅为 3.6\%；在全部数据集--主干组合上，宏平均降幅为 2.9\%，最大降幅为 24.9\%。它只增加约 0.7--4.3k 个参数，额外在线推理开销低于 8\,$\mu$s/sample。这些结果表明，TSF 能够把已有过程文档转化为跨主干和语义生成器的可测预测收益，同时保持轻量部署。
  Across multiple complex industrial forecasting and delayed soft-sensing tasks, TSF reduces MAE by 3.6\% on average. Across all dataset--backbone pairs, the macro-average reduction is 2.9\%, with a maximum reduction of 24.9\%. It adds only about 0.7--4.3k parameters, with less than 8\,$\mu$s/sample of additional online inference overhead. These results show that TSF turns existing process documents into measurable forecasting gains across backbones and semantic generators while remaining lightweight for deployment.
\end{abstract}


\section{Introduction}\label{sec:introduction}

% 过程工业依赖时序预测和软测量来估计难以在线测量的质量变量。这些预测直接影响安全裕度、产品质量、能耗和生产效率。软测量和冶金决策支持研究表明，数据驱动模型已经广泛用于工业信息系统~\cite{sun2021survey,zong2026llmdecision}。然而，真实部署常常先卡在数据条件上：高质量标签少，工况变化频繁，为不同场景重新训练模型或重建对齐流程的代价很高。与此同时，变量表和工艺文档通常已经存在，并记录变量名称、单位、物理含义、采样属性和工艺角色。这些文档本来就是训练前可用的知识来源，却常常没有真正进入预测过程。
Process industries rely on time-series forecasting and soft sensing to estimate quality variables that are hard to measure online. These predictions directly affect safety margins, product quality, energy use, and production efficiency. Work on soft sensing and metallurgy decision support shows that data-driven models are now widely used in industrial informatics~\cite{sun2021survey,zong2026llmdecision}. In real deployments, however, data conditions remain restrictive: high-quality labels are scarce, operating conditions change frequently, and retraining models or rebuilding alignment pipelines for each scenario is costly. Meanwhile, variable tables and process documents are usually already available. They record variable names, units, physical meanings, sampling attributes, and process roles. These documents are a natural source of prior knowledge, but they often remain outside the prediction process.

% 这一缺口在输入层最明显。多变量时序主干可以学习时间依赖，但输入映射通常只处理按列排列的数值窗口。工业任务里，变量不是匿名数值列，它们还携带单位、控制含义以及和目标变量的关系。尤其在样本有限时，模型需要在当前窗口中识别这些关系，而不是只记住每一列的数值形状。因此，更关键的是让输入变量与预测目标之间的关系在窗口进入主干之前就变得可见。
This gap is most visible at the input layer. Multivariate time-series backbones can learn temporal dependence, but their inputs are usually just column-ordered numerical windows. In industrial tasks, however, a variable is not an anonymous number. It also carries units, control meaning, and a relation to the target variable. When samples are limited, the model needs to recognize these relations in the current window rather than only memorize the numerical shape of each column. Therefore, the relation between input variables and the prediction target should be visible before the window enters the backbone.

\begin{figure*}[t]
  \centering
  \includegraphics[width=178mm]{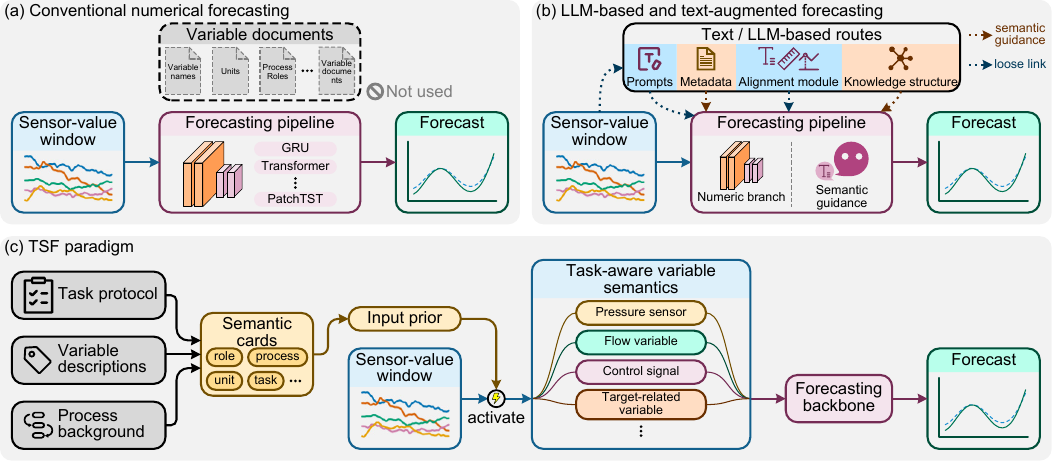}
  % 工业预测中变量语义的范式转变。(a)~传统数值预测只处理传感器值窗口，变量文档留在模型外部。(b)~LLM 和文本增强预测开始引入语义指导，但语义与当前窗口的联系仍可能偏松。(c)~TSF 先把任务协议和变量描述整理为任务感知语义，再由当前窗口在主干前激活这些语义。
  \caption{Paradigm transition for variable semantics in industrial forecasting. (a)~Conventional numerical forecasting only processes sensor-value windows, leaving variable documents outside the model. (b)~LLM-based and text-augmented forecasting introduces semantic guidance, but the link to the current numerical window can still remain loose. (c)~TSF first turns task protocols and variable descriptions into task-aware semantics, and then lets the current window activate these semantics before the backbone.}
  \label{fig:intro_tsf}
\end{figure*}

% 对大语言模型（LLM）和文本增强时间序列的研究表明，语言信息可以辅助时间序列建模~\cite{gruver2023llmtime,jin2024timellm,liu2024autotimes}。元数据和变量语义方法进一步表明，数据集描述和变量描述可以提供有用的预测先验。这些结果说明，变量描述不必只是被动文档。现有设计通常通过提示、元数据、对齐模块或知识结构等途径引入语义，但这些语义往往仍停留在预测管道的边缘。对于工业回归，更关键的是让当前观测值激活这些关系，并在每次预测中直接显示哪些变量关系正在参与。这张图说明了这种转变。
Studies on large language models (LLMs) and text-augmented time series show that language information can assist time-series modeling~\cite{gruver2023llmtime,jin2024timellm,liu2024autotimes}. Metadata and variable-semantics methods further show that dataset descriptions and variable descriptions can provide useful forecasting priors. These studies show that variable descriptions need not remain passive documentation. Existing designs often introduce semantics through prompts, metadata, alignment modules, or knowledge structures, but those semantics usually stay at the edge of the forecasting pipeline. For industrial regression, the key is to let the current numerical window activate these relations and reveal which variable relations are active in each prediction. \cref{fig:intro_tsf} illustrates this shift.

% 为此，我们提出 LLM-Guided Task-Semantic Field Factorization (TSF)。TSF 在训练前把任务协议、变量说明和工艺背景整理为任务感知的变量语义，并在预测时由当前数值窗口激活这些语义。这样，预测主干可以在任务场景和工况变化时，依托语义已被真正理解的输入变量实现动态适应。
To address this, we propose LLM-Guided Task-Semantic Field Factorization (TSF). Before training, TSF turns task protocols, variable descriptions, and process background into task-aware variable semantics, and the current numerical window activates these semantics at prediction time. In this way, the prediction backbone can adapt dynamically to changing task scenarios and operating conditions by relying on input variables whose semantics are properly understood.

% 主要贡献如下。
The main contributions are as follows.
\begin{itemize}
  \item
        % 我们提出 TSF，使用 LLM 的世界知识和推理来帮助模型识别输入变量与预测目标之间的语义逻辑关系。
        We propose TSF, which uses an LLM's world knowledge and reasoning to help models identify semantic-logical relations between input variables and the prediction target.
  \item
        % 我们引入任务语义场，使变量语义随任务定义和当前数值窗口被激活，并支持模型适应不同预测目标和工况变化。
        We introduce the task-semantic field, which activates variable semantics from the task definition and the current numerical window, and helps the model adapt to different prediction targets and operating shifts.
  \item
        % 在多个复杂工业预测和软测量任务上，TSF 在不同主干和语义生成器下提升性能，额外在线推理开销低于 8\,$\mu$s/sample。
        On multiple complex industrial forecasting and soft-sensing tasks, TSF improves performance across different backbones and semantic generators, with additional online inference overhead below 8\,$\mu$s/sample.
\end{itemize}
% 本文其余部分安排如下。\cref{sec:related_work} 回顾相关工作。\cref{sec:method} 介绍我们的方法。\cref{sec:experiments} 报告实验。\cref{sec:conclusion} 总结全文。
The remainder of this paper is organized as follows. \cref{sec:related_work} reviews related work. \cref{sec:method} presents the method. \cref{sec:experiments} reports the experiments. \cref{sec:conclusion} concludes the paper.

\section{Related Work}\label{sec:related_work}

% 数值主干与缺失的输入语义
\subsection{Numerical Backbones and Missing Input Semantics}\label{sec:2-1}

% 工业时序预测和软测量面临高质量标签稀缺、跨工况变化和部署限制。已有数值模型包括用于软测量的 LSTM 和 Transformer~\cite{yuan2020lstm,geng2022transformer}，以及用于材料压力预测的随机森林回归~\cite{zong2025hybridgrid}。
Industrial time-series forecasting and soft sensing face high-quality label scarcity, cross-regime variation, and deployment constraints. Existing numerical models include LSTM- and Transformer-based soft sensors~\cite{yuan2020lstm,geng2022transformer} and random forest regression for material-pressure prediction~\cite{zong2025hybridgrid}.
% 近期工业质量预测主要通过处理时序不确定性、融入过程结构和提高标签效率来改进数值建模。
Recent industrial quality-prediction studies improve numerical modeling mainly by addressing temporal uncertainty, incorporating process structure, and improving label efficiency.
% 时延感知方法联合估计变量时延和预测不确定性，从而提供质量变量的点预测和区间预测~\cite{yu2025delay}。
Delay-aware methods jointly estimate variable delays and predictive uncertainty to provide point and interval estimates of quality variables~\cite{yu2025delay}.
% 知识驱动方法通过网络模块和物理约束损失编码流场与浓度场之间的结构关系~\cite{wang2025knowledge}。
Knowledge-driven methods encode structural relations between flow and concentration fields through network modules and physics-constrained losses~\cite{wang2025knowledge}.
% 半监督软测量则结合性能驱动蒸馏和置信伪标签，以利用未标注的过程样本~\cite{yue2025distillation}。
Semi-supervised soft sensors combine performance-driven distillation with confident pseudo-labels to use unlabeled process samples~\cite{yue2025distillation}.
% 这些路线分别强化特定预测器的时序对齐、物理结构或监督信息，但尚未提供将现有变量文档转化为任务条件输入先验的通用接口。
These routes strengthen temporal alignment, physical structure, or supervision in specific predictors, but do not provide a common interface that turns existing variable documents into task-conditioned input priors.
% 变量到达模型之前，名称、单位、工艺角色以及与预测目标的关系往往已经不在输入中。
Before variables reach the model, their names, units, process roles, and relations to the prediction target are often absent from the input.
% TSF 因而把关注点放在主干之前的输入接口，使预测器在建模时间依赖之前看到保留任务语义的工业变量。
TSF therefore shifts attention to the input interface before the backbone, so that the forecaster can see industrial variables with their task semantics retained before modeling temporal dependence.

% 时序预测中的 LLM 接口
\subsection{LLM Interfaces for Time-Series Forecasting}\label{sec:2-2}

% 近期研究使用 LLM 补充数值窗口缺少的语义上下文，包括 prompt learning、直接生成、输入重编程、自回归建模以及时序--文本对齐~\cite{xue2024promptcast,gruver2023llmtime,jin2024timellm,zhou2023onefitsall,liu2024autotimes,liu2025timecma,sun2024test}。
Recent studies use LLMs to supplement the semantic context missing from numerical windows through prompt learning, direct generation, input reprogramming, autoregressive modeling, and time-series--text alignment~\cite{xue2024promptcast,gruver2023llmtime,jin2024timellm,zhou2023onefitsall,liu2024autotimes,liu2025timecma,sun2024test}.
% 更广泛的零样本基础模型应用给出相同启示。预训练能力只有结合目标任务和输入语境时才更可靠~\cite{su2025zero}。
Broader zero-shot foundation-model applications give a similar message. Pre-trained capability is more reliable when it is tied to the target task and input context~\cite{su2025zero}.
% 这种语义收益也伴随计算成本。受控研究表明，LLM 的额外开销并不总能转化为更好的预测结果~\cite{merrill2024llmts}。
This semantic benefit also comes with computational cost. Controlled studies show that the extra cost of an LLM does not always translate into better forecasting performance~\cite{merrill2024llmts}.
% TSF 遵循这种成本敏感的思路。它只在离线语义构造阶段使用 LLM，训练和推理由数值预测器完成。
TSF follows this cost-aware view. It uses the LLM only during offline semantic construction, while a numerical forecaster handles training and inference.

% 变量语义作为输入先验
\subsection{Variable Semantics as Input Priors}\label{sec:2-3}

% 近期基于 LLM 的工业智能沿多条路线引入领域上下文。重编程和适配器将退化知识或非结构化社会人口信息用于电池健康估计与负荷预测~\cite{chen2025dkilm,chen2026llmsa}。
Recent LLM-based industrial intelligence introduces domain context through several routes. Reprogramming and adapters incorporate degradation knowledge or unstructured socio-demographic information into battery health estimation and load forecasting~\cite{chen2025dkilm,chen2026llmsa}.
% 表示对齐和知识检索支持非平稳多变量预测，LLM 物理先验则辅助工业控制中的因果图构建~\cite{chu2026a2ra,yao2026causalllm}。
Representation alignment and knowledge retrieval support non-stationary multivariate forecasting, while LLM-derived physical priors assist causal-graph construction for industrial control~\cite{chu2026a2ra,yao2026causalllm}.
% 在工业故障诊断中，文本属性推理和知识图分别支持零样本识别、故障定位与处置建议~\cite{han2026zeroshot,liu2024kgllm}。
In industrial fault diagnosis, textual-attribute reasoning and knowledge graphs support zero-shot recognition, fault localization, and troubleshooting recommendations~\cite{han2026zeroshot,liu2024kgllm}.
% 这些路线表明，领域文本和语义关系能够改善工业建模，但通常在特定任务的训练或推理流程中保留 LLM 表示、检索或图推理。
These routes show that domain text and semantic relations can improve industrial modeling, but they usually retain LLM representations, retrieval, or graph reasoning in a task-specific training or inference pipeline.

% 与这些基于大模型的方法不同，TSF 仅在训练前使用 LLM，将任务协议和变量文档组织为变量--目标语义关系；在线预测器不包含 LLM、检索或图推理。
Unlike these LLM-based industrial approaches, TSF uses the LLM only before training to organize task protocols and variable documents into variable--target semantic relations; the online predictor contains no LLM, retrieval, or graph reasoning.
% 当前数值窗口在标准数值主干之前激活这些冻结语义，时序建模仍由原主干完成。
The current numerical window activates these frozen semantics before a standard numerical backbone, while temporal modeling remains with the original backbone.
% 与现有深度质量预测方法相比，TSF 不引入新的时序主干、物理约束损失或半监督目标，而是提供可跨任务和主干复用的语义输入接口。
Compared with existing deep quality-prediction methods, TSF does not introduce a new temporal backbone, physics-constrained loss, or semi-supervised objective. It provides a semantic input interface that can be reused across tasks and backbones.
% 因此，现有传感器表和变量文档可以成为输入侧先验，而无需成对语料或在线知识图维护。
Thus, existing sensor tables and variable documents can serve as input-side priors without paired corpora or online knowledge-graph maintenance.

\section{Method}\label{sec:method}

\subsection{Problem Setup and Semantic Directions}\label{sec:3-1}
% 我们首先定义每个预测样本，以及训练前固定的语义对象。
We first define each prediction sample and the semantic objects fixed before training.
% 每个样本由对象索引 $n$ 和预测时刻 $t$ 标识，窗口长度为 $L$，输入变量数为 $d$。
Each sample is identified by an instance index $n$ and a prediction time $t$. The window length is $L$, and the number of input variables is $d$.
% 对于该样本，令 $\mathbf{x}_{n,\ell}\in\mathbb{R}^{1\times d}$ 表示窗口内时刻 $\ell$ 的行向量，其中每列对应一个输入变量。
For this sample, let $\mathbf{x}_{n,\ell}\in\mathbb{R}^{1\times d}$ denote the row vector at time $\ell$ within the window. Each column corresponds to one input variable.
% 历史窗口 $X_{n,t}$ 由从 $t-L+1$ 到 $t$ 的 $L$ 个行向量堆叠而成。预测目标为 $\mathbf{y}_{n,t}\in\mathbb{R}^{r}$。
The history window $X_{n,t}$ is formed by stacking the $L$ rows from $t-L+1$ to $t$. The target is $\mathbf{y}_{n,t}\in\mathbb{R}^{r}$.
\begin{equation}
  X_{n,t}=
  \begin{bmatrix}
    \mathbf{x}_{n,t-L+1} \\
    \cdots               \\
    \mathbf{x}_{n,t}
  \end{bmatrix}
  \in\mathbb{R}^{L\times d},
  \qquad
  \mathbf{y}_{n,t}\in\mathbb{R}^{r}.
\end{equation}
% 其中 $r$ 是当前任务的输出维度。我们用 $\tilde{X}_{n,t}$ 表示通过训练集统计量归一化后的窗口，后续所有输入映射都作用在 $\tilde{X}_{n,t}$ 上。
Here, $r$ is the output dimension of the current task. We use $\tilde{X}_{n,t}$ for the window normalized with training-set statistics, and all input mappings below operate on $\tilde{X}_{n,t}$.
% 后续从该归一化窗口到主干输入的变换使用 \cref{tab:notation} 中汇总的符号。
The subsequent transformation from this normalized window to the backbone input uses the notation summarized in \cref{tab:notation}.

\begin{table}[t]
  \centering
  % TSF 使用的核心符号、维度和含义。
  \caption{Core symbols, dimensions, and meanings used in TSF.}
  \label{tab:notation}
  \setlength{\tabcolsep}{2pt}\scriptsize
  \begin{tabular}{ccp{0.68\linewidth}}
    \toprule
    Symbol & Dimension & Meaning \\
    \midrule
    $n,t,\ell,i$
      & index
      & Sample, prediction-time, within-window time, and variable indices. \\
    $L,d,r,k$
      & scalar
      & Window length and input, output, and semantic dimensions. \\
    $X_{n,t},\tilde{X}_{n,t}$
      & $L\times d$
      & Input window and its normalized form. \\
    $\mathbf{y}_{n,t},\hat{\mathbf{y}}_{n,t}$
      & $r$
      & Prediction target and model output. \\
    $\mathcal{T}$
      & --
      & Task protocol fixed before training. \\
    $\mathbf{v}_i$
      & $1\times k$
      & Semantic direction of the $i$-th input variable. \\
    $V$
      & $d\times k$
      & Frozen variable-semantic direction matrix. \\
    $S_{n,t}$
      & $L\times k$
      & Task-semantic field activated by the current window. \\
    $D$
      & $d\times d$
      & Diagonal scaling matrix for the raw-value path. \\
    $B$
      & $k\times d$
      & Learnable projection from the semantic field to the input dimension. \\
    $\mathbf{b}$
      & $d$
      & Bias shared across all time steps. \\
    $Z_{n,t}$
      & $L\times d$
      & Backbone input after TSF factorization. \\
    \bottomrule
  \end{tabular}
\end{table}

% TSF 的语义来自训练前固定的任务协议 $\mathcal{T}$。
TSF draws semantics from a task protocol $\mathcal{T}$ fixed before training.
% 我们定义 $\mathcal{T}=(\mathcal{B},\mathcal{G},\mathcal{I},\mathcal{C})$。其中 $\mathcal{B}$ 是过程背景，$\mathcal{G}$ 是目标和窗口设定，$\mathcal{I}$ 描述在线可得性与排除信息，$\mathcal{C}=(c_1,\ldots,c_d)$ 是按输入列顺序排列的变量元信息。
We define $\mathcal{T}=(\mathcal{B},\mathcal{G},\mathcal{I},\mathcal{C})$. Here, $\mathcal{B}$ is the process background, $\mathcal{G}$ is the target and window specification, and $\mathcal{I}$ describes information availability and excluded information. The variable metadata $\mathcal{C}=(c_1,\ldots,c_d)$ follows the input-column order.
\begin{equation}
  \mathcal{T}=(\mathcal{B},\mathcal{G},\mathcal{I},\mathcal{C}),
  \qquad
  \mathcal{C}=(c_1,\ldots,c_d).
\end{equation}
% 该协议记录数据集概况、可选领域上下文、变量名称、单位、采样属性、工艺角色和目标定义，并排除测试集统计、未来标签、预测误差和训练后模型状态。
This protocol records the dataset profile, optional domain context, variable names, units, sampling attributes, process roles, and target definitions. It excludes test-set statistics, future labels, prediction errors, and trained-model states.

% 离线语义生成器 $F_\phi$ 根据 $\mathcal{T}$ 自动生成语义卡，其中 $\phi$ 表示训练前确定的大语言模型和提示配置。
An offline semantic generator $F_\phi$ automatically maps $\mathcal{T}$ to semantic cards, where $\phi$ denotes the LLM and prompt configuration chosen before training.
% 生成结果 $(z_1,\ldots,z_d)$ 与输入变量顺序一致，其中 $z_i$ 是第 $i$ 个变量的语义卡。
The output $(z_1,\ldots,z_d)$ follows the same order as the input variables, where $z_i$ is the semantic card of the $i$-th variable.
\begin{equation}
  (z_1,\ldots,z_d)=F_\phi(\mathcal{T}).
\end{equation}
% 每张语义卡 $z_i$ 描述第 $i$ 个变量在当前任务中的名称、单位、物理含义、过程角色、信息角色、时间关系和主要耦合关系。
Each semantic card $z_i$ describes the name, unit, physical meaning, process role, information role, temporal relation, and main coupling relations of the $i$-th variable in the current task.

% 在文本渲染和 embedding 之前，我们依据训练前可用的结构化变量表或数据字典，对每张候选语义卡进行冻结前验证。验证检查必填字段、标签唯一性、输入列顺序、单位、物理含义和缩写解释。若候选字段与来源记录冲突，则以结构化记录为准；仅在现有记录无法消除歧义时才进行人工确认。该过程修正可核验字段，同时保留与任务协议一致的 LLM 生成物理关系和变量耦合。验证通过的语义卡随后被渲染和嵌入，并用于构建冻结矩阵 $V$。\cref{tab:semantic_card_examples} 给出了 IndPenSim 中两个原始缩写标签的验证示例。
Before text rendering and embedding, each candidate semantic card undergoes pre-freezing validation against structured variable tables or data dictionaries available before training. The checks cover required fields, tag uniqueness, input-column order, units, physical meanings, and abbreviation expansions. When a candidate field conflicts with a source record, the structured record takes precedence; manual confirmation is used only when the available records cannot resolve the ambiguity. This process corrects verifiable fields while retaining LLM-generated physical relations and variable couplings that agree with the task protocol. Validated cards are then rendered, embedded, and used to construct the frozen matrix $V$. \cref{tab:semantic_card_examples} gives two validation examples for raw abbreviated tags in IndPenSim.

\begin{table*}[t]
  \centering
  % IndPenSim 中原始变量标签的冻结前验证示例。结构化变量记录用于识别并纠正候选语义卡中的缩写解释、过程角色和单位冲突。
  \caption{Pre-freezing validation examples for raw variable tags in IndPenSim. Structured variable records identify and correct conflicts in abbreviation expansion, process role, and unit.}
  \label{tab:semantic_card_examples}
  \setlength{\tabcolsep}{2pt}\scriptsize
  {
    \begin{tabular}{p{0.05\linewidth}p{0.16\linewidth}p{0.16\linewidth}p{0.28\linewidth}p{0.29\linewidth}}
      \toprule
      Raw tag\newline(position) & Structured record & Candidate card excerpt & Validation & Validated card excerpt \\
      \midrule
      \texttt{wfi}\newline(Col. 9) & Water-for-Injection Flow; L/h; water-feed control input & Water Flow Indicator; L/h; flow observation & Expansion and process role conflict with the data dictionary; the structured record takes precedence. & Water-for-Injection Flow; L/h; control input affecting broth dilution and volume. Corrected and accepted. \\
      \addlinespace[2pt]
      \texttt{our}\newline(Col. 20) & Oxygen Uptake Rate; g/min; derived metabolic variable & Oxygen Uptake Rate; g/h; derived metabolic variable & Unit conflicts with the variable table; the recorded unit takes precedence. & Oxygen Uptake Rate; g/min; derived variable reflecting microbial respiration. Corrected and accepted. \\
      \bottomrule
    \end{tabular}
  }
\end{table*}

% 这些语义卡随后被整理为变量语义方向。
The semantic cards are then converted into variable-semantic directions.

\begin{figure*}[t]
  \centering
  \includegraphics[width=178mm]{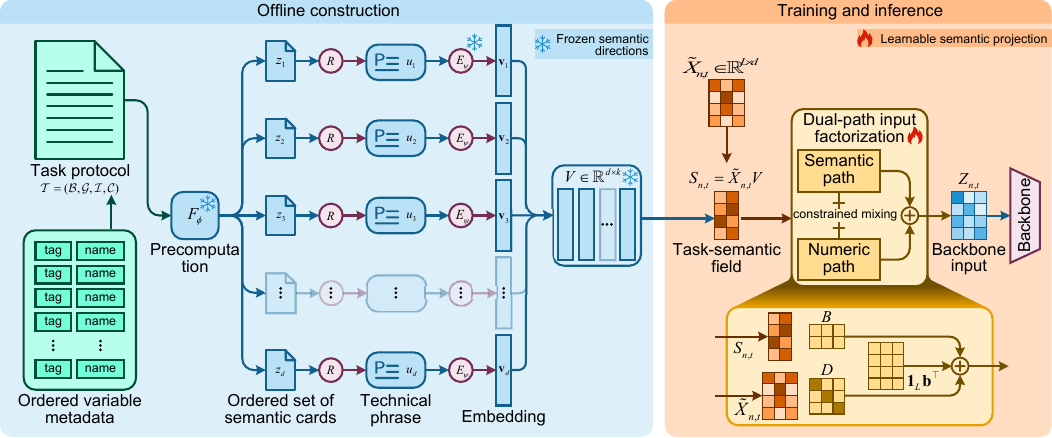}
  % TSF 工作流。离线构建阶段将任务协议和有序变量元信息映射为语义卡、渲染短语、嵌入向量和冻结的变量语义方向矩阵 $V$。训练和推理阶段中，归一化窗口激活任务语义场 $S=\tilde{X}V$。受限双路径适配层将语义场、原始数值路径和共享偏置合成为主干输入 $Z$；插图展开该因子化过程。
  \caption{TSF workflow. Offline construction maps the task protocol and ordered variable metadata to semantic cards, rendered phrases, embeddings, and the frozen variable-semantic direction matrix $V$. During training and inference, the normalized window activates the task-semantic field $S=\tilde{X}V$. The constrained dual-path adapter combines the semantic field, raw-value path, and shared bias into the backbone input $Z$; the inset expands this factorization.}
  \label{fig:tsf_method_overview}
\end{figure*}

% 确定性文本渲染器 $R$ 将语义卡 $z_i$ 转为紧凑英文短语 $u_i$；embedding 模型 $E_\psi$ 再将 $u_i$ 映射为向量，其中 $\psi$ 是该 embedding 模型的冻结参数。
The deterministic text renderer $R$ converts the semantic card $z_i$ into a compact English phrase $u_i$. The embedding model $E_\psi$ then maps $u_i$ to a vector, where $\psi$ denotes the frozen parameters of this embedding model.
% 归一化后的行向量 $\mathbf{v}_i\in\mathbb{R}^{1\times k}$ 定义为第 $i$ 个变量的变量语义方向，$k$ 是语义方向维度。
The normalized row vector $\mathbf{v}_i\in\mathbb{R}^{1\times k}$ defines the variable-semantic direction of the $i$-th variable, and $k$ is the semantic-direction dimension.
\begin{equation}
  u_i=R(z_i),
  \qquad
  \mathbf{v}_i=
  \frac{E_\psi(u_i)}
  {\left\lVert E_\psi(u_i)\right\rVert_2}
  \in\mathbb{R}^{1\times k}.
\end{equation}
% 将所有变量语义方向按输入列顺序堆叠，得到变量语义方向矩阵 $V$。
Stacking all variable-semantic directions in input-column order gives the variable-semantic direction matrix $V$.
\begin{equation}
  V=
  \begin{bmatrix}
    \mathbf{v}_1 \\
    \cdots       \\
    \mathbf{v}_d
  \end{bmatrix}
  \in\mathbb{R}^{d\times k}.
\end{equation}
% $V$ 在训练开始前已经确定，并在训练、验证和测试中保持不变。
After construction, $V$ remains unchanged during training, validation, and testing.
% \cref{fig:tsf_method_overview} 总结了离线语义构建以及训练和推理阶段的计算路径。
\cref{fig:tsf_method_overview} summarizes offline semantic construction and the computation path used in training and inference.

\subsection{Task-Semantic Field Factorization}\label{sec:3-2}
% 在得到变量语义方向矩阵 $V$ 后，TSF 用当前数值状态来激活它。
With the variable-semantic direction matrix $V$ constructed, TSF activates it using the current numerical state.
% 对窗口内任意时刻 $\ell$，$\tilde{\mathbf{x}}_{n,\ell}\in\mathbb{R}^{1\times d}$ 表示归一化输入行向量，$\tilde{x}_{n,\ell,i}$ 是其中第 $i$ 个元素。
For any time index $\ell$ in the window, $\tilde{\mathbf{x}}_{n,\ell}\in\mathbb{R}^{1\times d}$ denotes the normalized input row, and $\tilde{x}_{n,\ell,i}$ is its $i$-th element.
% 对应的逐时刻任务语义场向量 $\mathbf{s}_{n,\ell}$ 是各变量语义方向的数值加权和。
The corresponding row-wise task-semantic field vector $\mathbf{s}_{n,\ell}$ is a value-weighted sum of the variable-semantic directions.
\begin{equation}
  \mathbf{s}_{n,\ell}
  =
  \tilde{\mathbf{x}}_{n,\ell}V
  =
  \sum_{i=1}^{d}\tilde{x}_{n,\ell,i}\mathbf{v}_i
  \in\mathbb{R}^{1\times k}.
\end{equation}
% 将所有 $L$ 个时刻的 $\mathbf{s}_{n,\ell}$ 按时间顺序堆叠，得到窗口级任务语义场 $S_{n,t}\in\mathbb{R}^{L\times k}$。
Stacking all $L$ row-wise vectors $\mathbf{s}_{n,\ell}$ in temporal order gives the window-level task-semantic field $S_{n,t}\in\mathbb{R}^{L\times k}$.
\begin{equation}
  S_{n,t}=\tilde{X}_{n,t}V.
\end{equation}
% 尽管 $V$ 在训练前固定，$S_{n,t}$ 仍随当前数值窗口变化。在每个时刻，各变量的当前数值分别缩放对应的语义方向。这些加权方向共同表示当前过程状态的任务语义。该线性运算以较低开销在主干前加入语义信息，窗口内的时间依赖仍由主干建模。
Although $V$ is fixed before training, $S_{n,t}$ changes with the current numerical window. At each time step, the current value of each variable scales its semantic direction. The weighted directions jointly represent the task semantics of the current process state. This linear operation adds semantic information before the backbone with low overhead. The backbone continues to model temporal dependencies within the window.

\begin{figure}[t]
  \centering
  \includegraphics[width=88mm]{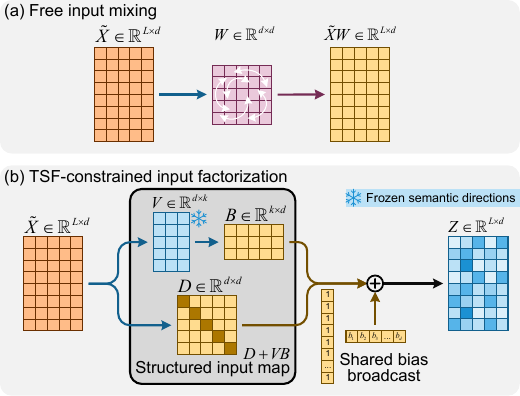}
  % 自由输入混合与 TSF 受限输入因子化。(a)~传统主干前输入层学习任意稠密矩阵 $W$，并形成 $\tilde{X}W$。(b)~TSF 使用 $Z=\tilde{X}(D+VB)+\mathbf{1}_{L}\mathbf{b}^{\top}$，其中 $D$ 保留逐变量数值通道，$VB$ 通过冻结变量语义方向 $V$ 和可学习投影 $B$ 约束跨变量混合。
  \caption{Free input mixing versus TSF-constrained input factorization. (a)~A conventional pre-backbone layer learns an arbitrary dense matrix $W$ and forms $\tilde{X}W$. (b)~TSF uses $Z=\tilde{X}(D+VB)+\mathbf{1}_{L}\mathbf{b}^{\top}$, where $D$ preserves per-variable numerical channels and $VB$ constrains cross-variable mixing through the frozen variable-semantic directions $V$ and the learnable projection $B$.}
  \label{fig:tsf_factorization}
\end{figure}

% TSF 随后在主干前使用双路径输入适配器，以同时保留原始数值通道并引入语义场投影。
TSF then uses a dual-path input adapter before the backbone to preserve the raw-value channels and introduce the semantic-field projection.
% 在原始数值路径中，对角矩阵 $D\in\mathbb{R}^{d\times d}$ 保留逐变量尺度，其中 $\boldsymbol{\alpha}\in\mathbb{R}^{d}$ 是可学习的逐变量尺度向量。
In the raw-value path, the diagonal matrix $D\in\mathbb{R}^{d\times d}$ preserves per-variable scaling, where $\boldsymbol{\alpha}\in\mathbb{R}^{d}$ is a learnable per-variable scale vector.
% 语义场路径使用可学习投影 $B\in\mathbb{R}^{k\times d}$；$\mathbf{b}\in\mathbb{R}^{d}$ 是偏置向量，$\mathbf{1}_{L}\in\mathbb{R}^{L}$ 用于把该偏置复制到窗口的每个时刻。
The semantic-field path uses a learnable projection $B\in\mathbb{R}^{k\times d}$. The vector $\mathbf{b}\in\mathbb{R}^{d}$ is the bias, and $\mathbf{1}_{L}\in\mathbb{R}^{L}$ copies this bias to every time step in the window.
\begin{equation}
  \begin{gathered}
    D=\operatorname{diag}(\boldsymbol{\alpha}),\quad
    \boldsymbol{\alpha}\in\mathbb{R}^{d},\quad
    B\in\mathbb{R}^{k\times d},\\
    \mathbf{b}\in\mathbb{R}^{d},\quad
    \mathbf{1}_{L}\in\mathbb{R}^{L}.
  \end{gathered}
\end{equation}
% 主干输入表示 $Z_{n,t}\in\mathbb{R}^{L\times d}$ 与原始输入窗口具有相同的时间长度和变量维度。
The backbone input representation $Z_{n,t}\in\mathbb{R}^{L\times d}$ has the same time length and variable dimension as the original input window.
\begin{equation}
  Z_{n,t}
  =
  \tilde{X}_{n,t}D
  +
  S_{n,t}B
  +
  \mathbf{1}_{L}\mathbf{b}^{\top}.
\end{equation}
% 从左到右，该式的三项分别对应原始数值路径、语义场投影路径和逐时刻共享的偏置项。
From left to right, the three terms correspond to the raw-value path, the semantic-field projection path, and the bias shared across time steps.
% 将 $S_{n,t}=\tilde{X}_{n,t}V$ 代入后，可得到主干前输入映射的受限结构。
Substituting $S_{n,t}=\tilde{X}_{n,t}V$ gives the constrained structure of the pre-backbone input mapping.
\begin{equation}
  Z_{n,t}
  =
  \tilde{X}_{n,t}(D+VB)
  +
  \mathbf{1}_{L}\mathbf{b}^{\top}.
\end{equation}
% 该等价式表明，跨变量混合来自冻结变量语义方向 $V$ 和可学习投影 $B$ 的乘积 $VB$，而对角项 $D$ 保留每个变量的数值通道。
This equivalent form shows that cross-variable mixing comes from the product $VB$ of the frozen variable-semantic directions $V$ and the learnable projection $B$, while the diagonal term $D$ preserves each variable's numerical channel.
% \cref{fig:tsf_factorization} 将这种受限映射与自由输入混合进行对比。自由层学习任意矩阵 $W\in\mathbb{R}^{d\times d}$，而 TSF 将映射限制为对角残差 $D$ 和语义约束乘积 $VB$。
\cref{fig:tsf_factorization} compares this constrained map with free input mixing. A free layer learns an arbitrary matrix $W\in\mathbb{R}^{d\times d}$, whereas TSF restricts the map to the diagonal residual $D$ and the semantically constrained product $VB$.

\subsection{Training and Inference}\label{sec:3-3}
% 输入适配器形成 $Z_{n,t}$ 后，TSF 可以与任意接收 $L\times d$ 窗口表示的时序主干连接。
After the input adapter forms $Z_{n,t}$, TSF can be connected to any time-series backbone that accepts an $L\times d$ window representation.
% 令 $G_\theta$ 表示该主干及其预测头，$\theta$ 包含其中的可学习参数，预测结果写为 $\hat{\mathbf{y}}_{n,t}$。
Let $G_\theta$ denote the backbone together with its prediction head. Its learnable parameters are included in $\theta$, and its prediction is written as $\hat{\mathbf{y}}_{n,t}$.
\begin{equation}
  \hat{\mathbf{y}}_{n,t}=G_\theta(Z_{n,t}).
\end{equation}
% 训练样本索引集合为 $\mathcal{I}_{\mathrm{train}}$，监督回归损失为 $\mathcal{L}$。
The training index set is $\mathcal{I}_{\mathrm{train}}$, and the supervised regression loss is $\mathcal{L}$.
% 需要学习的参数集合为 $\Theta=\{\theta,B,\boldsymbol{\alpha},\mathbf{b}\}$，其中 $\boldsymbol{\alpha}$ 通过 $D=\operatorname{diag}(\boldsymbol{\alpha})$ 决定原始数值路径。
The trainable parameter set is $\Theta=\{\theta,B,\boldsymbol{\alpha},\mathbf{b}\}$, where $\boldsymbol{\alpha}$ determines the raw-value path through $D=\operatorname{diag}(\boldsymbol{\alpha})$.
% 经验风险最小化目标为
The empirical risk minimization objective is
\begin{equation}
  \Theta^\ast
  =
  \arg\min_{\Theta}
  \frac{1}{|\mathcal{I}_{\mathrm{train}}|}
  \sum_{(n,t)\in\mathcal{I}_{\mathrm{train}}}
  \mathcal{L}
  \big(
  G_\theta(Z_{n,t}),
  \mathbf{y}_{n,t}
  \big),
\end{equation}
% 训练过程中，$F_\phi$、$R$、$E_\psi$ 和 $V$ 保持不变；可学习部分只包括主干参数、语义场投影、原始数值路径尺度和偏置。
During training, $F_\phi$, $R$, $E_\psi$, and $V$ remain unchanged. Only the backbone parameters, semantic-field projection, raw-value path scales, and bias are learned.
% 推理时，TSF 复用同一个 $V$，并依次计算 $\tilde{X}_{n,t}$、$S_{n,t}$、$Z_{n,t}$ 和 $G_\theta(Z_{n,t})$。
At inference, TSF reuses the same $V$ and computes $\tilde{X}_{n,t}$, $S_{n,t}$, $Z_{n,t}$, and $G_\theta(Z_{n,t})$ in order.
% 测试集的唯一作用是最终评估；归一化统计估计、语义卡生成、embedding 生成和模型选择均不使用测试集。
The only role of the test set is final evaluation. Normalization-statistic estimation, semantic-card generation, embedding generation, and model selection are completed without using the test set.
% 在公平比较中，baseline 直接接收 $\tilde{X}_{n,t}$，TSF 接收 $Z_{n,t}$；二者共享相同的数据划分、归一化策略、主干结构和训练预算。
For a fair comparison, the baseline receives $\tilde{X}_{n,t}$ directly, while TSF receives $Z_{n,t}$. Both use the same data split, normalization strategy, backbone architecture, and training budget.
% 因此，TSF 引入的唯一模型变化是主干前的语义输入因子化。
Thus, the only model change introduced by TSF is semantic input factorization before the backbone.

\section{Experiments}\label{sec:experiments}

% 本节在四个工业预测和软测量任务上评估 TSF。
This section evaluates TSF on four industrial forecasting and soft-sensing tasks.

\subsection{Experimental Settings}\label{sec:4-1}

% 所有方法在同一实现中按照同一训练协议评估。对于每个数据集-主干组合，Base 与 +TSF 共享数据划分、归一化、主干配置、训练预算和模型选择规则。测试集不参与调参。
All methods are evaluated in the same implementation and under the same training protocol. For each dataset--backbone pair, Base and +TSF share the data split, normalization, backbone configuration, training budget, and model-selection rule. The test set is never used for tuning.
% 所有实验在一台配备 Intel Xeon 8470Q CPU、90\,GB RAM 和单张 NVIDIA RTX 5090 GPU（32\,GB）的工作站上运行。实现基于 Python~3.14、PyTorch~2.11.0 和 CUDA~13.0。所有结果均采用固定的五个不同随机种子。
All experiments were run on a workstation with an Intel Xeon 8470Q CPU, 90\,GB RAM, and a single NVIDIA RTX 5090 GPU (32\,GB). The implementation uses Python~3.14, PyTorch~2.11.0, and CUDA~13.0. All results use a fixed set of five distinct random seeds.
% 为便于复现，源代码已在公开 GitHub 仓库中提供。
For reproducibility, the source code is available at \href{https://github.com/mituan-ai/TSF_open}{https://github.com/mituan-ai/TSF\_open}.
% \cref{tab:experiment_config} 总结了实验配置。
\cref{tab:experiment_config} summarizes the experimental settings.

\begin{table}[t]
  \centering
  % 所有实验的通用训练、评估和 TSF 语义构造设置。对于每个数据集-主干组合，Base 与 +TSF 共享相同划分、优化器、训练预算和模型选择规则。
  \caption{Common training, evaluation, and TSF semantic construction settings for all experiments. For each dataset--backbone pair, Base and +TSF share the same split, optimizer, training budget, and model-selection rule.}
  \label{tab:experiment_config}
  \setlength{\tabcolsep}{4pt}\scriptsize
  \begin{tabular}{lll}
    \toprule
    Name                    & \multicolumn{2}{l}{Setting}                                    \\
    \midrule
    Optimizer               & \multicolumn{2}{l}{AdamW}                                      \\
    Learning rate           & \multicolumn{2}{l}{10\textsuperscript{-3}}                     \\
    Loss                    & \multicolumn{2}{l}{MSE}                                        \\
    Max epochs              & \multicolumn{2}{l}{300}                                        \\
    Early stopping          & patience                                   & 40                \\
    \multirow[t]{3}{*}{LR scheduler}
                            & type                                       & ReduceLROnPlateau \\
                            & decay factor                               & 0.5               \\
                            & patience                                   & 15                \\
    Weight decay            & \multicolumn{2}{l}{10\textsuperscript{-4}}                     \\
    Gradient clipping       & \multicolumn{2}{l}{1.0}                                        \\
    Batch size              & \multicolumn{2}{l}{64}                                         \\
    Semantic dimension      & $k$                                        & 128               \\
    Semantic-card generator & \multicolumn{2}{l}{DeepSeek-v4-pro (max)}                      \\
    Embedding model         & \multicolumn{2}{l}{text-embedding-v4}                          \\
    \bottomrule
  \end{tabular}
\end{table}

% 报告的时间是测试集上摊销到单个样本的在线推理时间。对每个训练完成的模型，测试批次预先加载到 GPU。先进行 10 次完整测试集预热，再对 80 次完整测试集前向传播计时，并在计时前后同步 CUDA。计时包含 TSF 适配器和预测模型的前向计算，但不包含数据加载、后处理以及离线 LLM 语义卡和 embedding 生成。
The reported time is the amortized online inference time per test sample. For each trained model, the test batches are preloaded onto the GPU. We use 10 full test-set passes for warm-up and time a further 80 passes, with CUDA synchronized before and after timing. The measurement includes the forward computation of the TSF adapter and forecasting model, but excludes data loading, post-processing, and offline LLM semantic-card and embedding generation.
% 主结果报告平均绝对误差（MAE）、均方根误差（RMSE）、决定系数（$R^2$）和单样本推理时间（$\mu$s/sample）。参数量和峰值 RSS 内存在在线成本分析中报告。
The main results report mean absolute error (MAE), root mean squared error (RMSE), coefficient of determination ($R^2$), and per-sample inference time ($\mu$s/sample). Parameter count and peak RSS memory are reported in the online-cost analysis.
% 比较覆盖八种常用时序主干。它们包括 GRU 和 LSTM~\cite{cho2014gru,hochreiter1997lstm}、Transformer 和 Informer~\cite{vaswani2017transformer,zhou2021informer}、Mamba 和 iTransformer~\cite{gu2024mamba,liu2024itransformer}，以及 PatchTST 和 ModernTCN~\cite{nie2023patchtst,luo2024moderntcn}。在每个数据集和主干下，Base 与 +TSF 仅在输入适配器上不同。
The comparison covers eight common time-series backbones. They include GRU and LSTM~\cite{cho2014gru,hochreiter1997lstm}, Transformer and Informer~\cite{vaswani2017transformer,zhou2021informer}, Mamba and iTransformer~\cite{gu2024mamba,liu2024itransformer}, and PatchTST and ModernTCN~\cite{nie2023patchtst,luo2024moderntcn}. For each dataset and backbone, Base and +TSF differ only in the input adapter.

\subsection{Datasets}\label{sec:4-2}

% 实验使用四个数据集，分别来自钢铁冶金、尾矿浓密脱水、补料分批发酵和 Tennessee Eastman 化工过程。它们覆盖未来温度预测、当前底流浓度软测量、化验时刻青霉素浓度估计和延迟产品组分软测量。数据来源包含两个私有现场数据集和两个公开工业仿真基准。\cref{tab:dataset_protocol} 给出每个数据集的变量、目标和预定义划分。
The experiments use four datasets from steel metallurgy, tailings thickening dewatering, fed-batch fermentation, and the Tennessee Eastman chemical process. They cover future temperature forecasting, current underflow-concentration soft sensing, assay-time penicillin-concentration estimation, and delayed product-composition soft sensing. The sources include two private plant datasets and two public industrial simulation benchmarks. \cref{tab:dataset_protocol} gives the variables, targets, and predefined splits for each dataset.
% 所有训练、验证和测试划分在训练前固定，并在过程、场景或批次级别完成。训练数据主要来自常规或近域运行，测试数据包含完整保留过程、近域对照以及未训练过的控制或故障工况。该协议避免使用更容易泄漏的随机窗口划分。
All training, validation, and test splits are fixed before training at the process, scenario, or batch level. Training data mainly come from routine or near-domain operation, while test data contain complete held-out processes, near-domain references, and unseen control or fault regimes. This protocol avoids the easier random-window split, which can leak information through the same run or adjacent windows.

\begin{table}[t]
  \centering
  % TSF 实验使用的数据集协议。表中列出每个过程的输入变量、采样间隔、窗口长度、预测目标以及固定的训练、验证和测试划分。
  \caption{Dataset protocols used in the TSF experiments. The table lists input variables, sampling intervals, window lengths, prediction targets, and fixed train/validation/test splits for each process.}
  \label{tab:dataset_protocol}
  \setlength{\tabcolsep}{2pt}\scriptsize
  \begin{tabular}{p{0.12\linewidth}p{0.6\linewidth}rl}
    \toprule
    Item   & \multicolumn{3}{l}{Description}                                                       \\
    \midrule
    \multicolumn{4}{l}{\textit{Ladle Preheating}}                                                  \\
    \multirow[t]{3}{*}{Input}
           & Variables                                                          & 14     &         \\
           & Sampling interval                                                  & 5      & minutes \\
           & Input window                                                       & 60     & steps   \\
    Target & \multicolumn{3}{l}{Ladle temperature over the next 5 steps}                           \\
    \multirow[t]{3}{*}{Splits}
           & 24 training processes                                              & 15,236 & samples \\
           & 5 validation processes                                             & 2,500  & samples \\
           & 3 held-out test processes                                          & 2,104  & samples \\
    \midrule
    \multicolumn{4}{l}{\textit{Thickener Dewatering}}                                              \\
    \multirow[t]{3}{*}{Input}
           & Variables                                                          & 5      &         \\
           & Sampling interval                                                  & 1      & minute  \\
           & Input window                                                       & 30     & steps   \\
    Target & \multicolumn{3}{l}{Underflow concentration at the current step}                       \\
    \multirow[t]{4}{*}{Splits}
           & 5 training mild-jitter scenarios                                   & 1,355  & samples \\
           & 1 validation mild-jitter scenario                                  & 271    & samples \\
           & 1 near-domain test scenario                                        & 271    & samples \\
           & 12 far-domain test scenarios                                       & 3,252  & samples \\
    \midrule
    \multicolumn{4}{l}{\textit{IndPenSim}}                                                         \\
    \multirow[t]{3}{*}{Input}
           & Variables                                                          & 23     &         \\
           & Sampling interval                                                  & 12     & minutes \\
           & Input window                                                       & 120    & steps   \\
    Target & \multicolumn{3}{l}{Assay-time penicillin concentration}                               \\
    \multirow[t]{5}{*}{Splits}
           & 24 training recipe-controlled batches                              & 451    & samples \\
           & 5 validation recipe-controlled batches                             & 90     & samples \\
           & 1 same-family recipe-controlled test batch                         & 19     & samples \\
           & 2 near-domain operator-controlled test batches                     & 38     & samples \\
           & 12 far-domain advanced process control (APC) or fault test batches & 220    & samples \\
    \midrule
    \multicolumn{4}{l}{\textit{Tennessee Eastman Process}}                                       \\
    \multirow[t]{3}{*}{Input}
           & Variables                                                          & 33     &         \\
           & Sampling interval                                                  & 3      & minutes \\
           & Input window                                                       & 20     & steps   \\
    Target & \multicolumn{3}{l}{Delayed product-stream component G concentration}                 \\
    \multirow[t]{5}{*}{Splits}
           & 30 training trajectories                                           & 16,380 & samples \\
           & 5 validation trajectories                                          & 2,730  & samples \\
           & 15 same-family test trajectories                                   & 1,365  & samples \\
           & 15 near-domain test trajectories                                   & 6,825  & samples \\
           & 15 far-domain test trajectories                                    & 20,475 & samples \\
    \bottomrule
  \end{tabular}
\end{table}

\begin{figure*}[t]
  \centering
  \includegraphics[width=173mm]{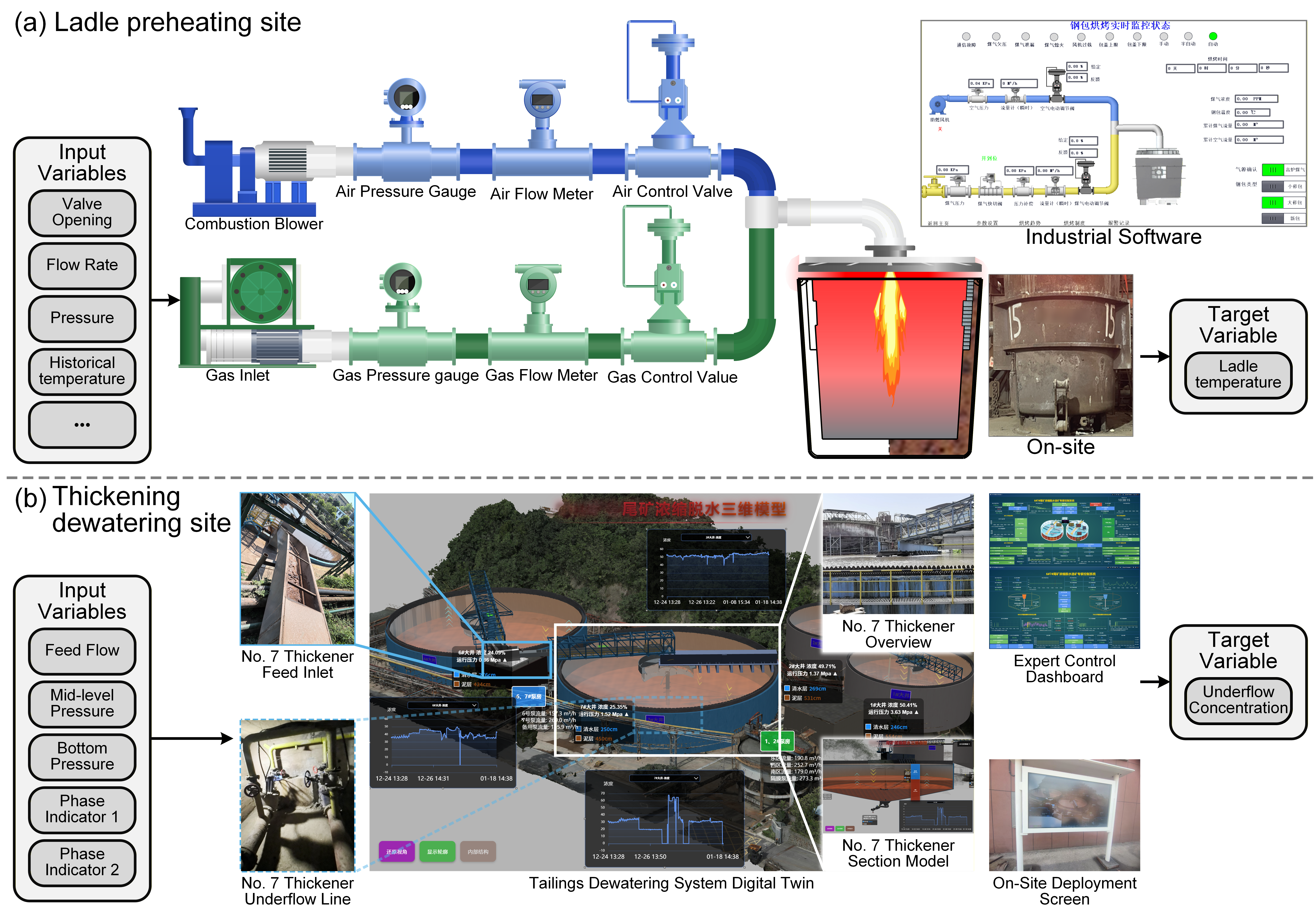}
  % 两个工业现场过程示例。(a)~中国马鞍山某炼钢厂钢包烘烤现场。(b)~中国南京某厂尾矿浓密脱水现场。
  \caption{On-site views of the two plant processes. (a)~Ladle preheating at a steel plant in Maanshan, China. (b)~Tailings thickening dewatering at a plant in Nanjing, China.}
  \label{fig:industrial_sites}
\end{figure*}

% 钢包预热数据集来自私有钢铁冶金过程，该过程也用于变量钢包烘烤趋势预测研究~\cite{zong2025metacontrastive}。该任务根据燃气侧、空气侧、燃烧状态和历史温度预测未来钢包内部温度。划分按完整预热过程完成；每个测试过程都是独立运行，但没有显式场景标签或外部工况描述。
The ladle preheating dataset comes from a private steel-metallurgy process that has also been used in variable ladle-preheating trend prediction~\cite{zong2025metacontrastive}. The task predicts future ladle inner temperature from fuel-gas, air-side, combustion-state, and historical-temperature variables. The split is made by complete preheating processes; each test process is an independent run, but no explicit scenario labels or external regime descriptions are provided.
% 浓密脱水数据集来自尾矿浓密脱水过程~\cite{lin2025expertaugmented}。该任务根据进料、压力和运行阶段变量估计当前底流浓度。训练和验证数据来自 mild-jitter 场景，测试集包含近域对照和远域工况迁移。
The thickener dewatering dataset comes from a tailings thickening dewatering process~\cite{lin2025expertaugmented}. The task estimates current underflow concentration from feed, pressure, and operating-phase variables. Training and validation data come from mild-jitter scenarios, while the test set includes a near-domain reference and far-domain operating shifts.
% IndPenSim 数据集来自公开的 IndPenSim V3 青霉素补料分批发酵模拟器~\cite{goldrick2015development}。该任务根据化验时刻之前的在线过程窗口估计离线青霉素浓度。训练批次采用 recipe-controlled 运行，测试批次覆盖同族配方、操作员控制、advanced process control (APC) 运行和故障条件。
The IndPenSim dataset is drawn from the public IndPenSim V3 penicillin fed-batch fermentation simulator~\cite{goldrick2015development}. The task estimates offline penicillin concentration from the online process window before each assay time. Training batches use recipe-controlled operation, while test batches cover same-family recipes, operator control, advanced process control (APC) operation, and fault conditions.
% Tennessee Eastman Process (TEP) 数据集来自公开的扩展仿真数据~\cite{rieth2017tep}。该任务根据 22 个连续过程测量和 11 个操纵变量的 60 分钟窗口，估计延迟报告的产品流组分 G。划分同时隔离仿真轨迹和工况，测试集包含同族运行、近域变化和训练中未见的远域扰动。工况标签仅用于划分和结果解释，不作为模型输入。
The Tennessee Eastman Process (TEP) dataset is drawn from publicly available extended simulation data~\cite{rieth2017tep}. The task estimates the delayed product-stream component G from a 60-min window of 22 continuous process measurements and 11 manipulated variables. The split isolates both simulation trajectories and operating conditions, and the test set contains same-family operation, near-domain variation, and unseen far-domain disturbances. Condition labels are used only for splitting and result interpretation, not as model inputs.

\subsection{Main Results}\label{sec:4-3}

% \cref{tab:main_ladle}、\cref{tab:main_thickener}、\cref{tab:main_indpensim} 和 \cref{tab:main_tep} 比较 Base 与 +TSF 在四个测试集上的预测性能。这些测试集包含 \cref{tab:dataset_protocol} 中定义的完整保留过程和场景迁移样本。
\cref{tab:main_ladle}, \cref{tab:main_thickener}, \cref{tab:main_indpensim}, and \cref{tab:main_tep} compare the forecasting performance of Base and +TSF on the four test sets. These sets include the held-out processes and scenario-shifted samples defined in \cref{tab:dataset_protocol}.
% 对每个数据集--主干组合的五次运行 MAE 采用双侧配对 $t$ 检验，$^{*}$ 表示 $p<0.05$。
For each dataset--backbone pair, a two-sided paired $t$-test is applied to the five-run MAE values, with $^{*}$ denoting $p<0.05$.

\begin{table}[t]
  \centering
  % 钢包预热数据集上的定量对比。$^{*}$ 表示 MAE 的 $p<0.05$。
  \caption{Quantitative comparison on the ladle preheating dataset. $^{*}$ denotes $p<0.05$ for MAE.}
  \label{tab:main_ladle}
  \setlength{\tabcolsep}{2pt}\scriptsize
  \begin{tabular}{llcccr}
    \toprule
    Backbone                      & Setting & MAE$\downarrow$             & RMSE$\downarrow$            & R$^2$ $\uparrow$            & \multicolumn{1}{c}{\makecell{Inference Time\\($\mu$s/sample)$\downarrow$}} \\
    \midrule
    \multirow{2}{*}{GRU$^{*}$}    & Base    & 0.018{\tiny\,\textpm0.0001} & 0.031{\tiny\,\textpm0.0002} & 0.991{\tiny\,\textpm0.0001} & 3.52{\tiny\,\textpm0.03}  \\
                                  & +TSF    & 0.016{\tiny\,\textpm0.0008} & 0.030{\tiny\,\textpm0.0009} & 0.991{\tiny\,\textpm0.0005} & 3.73{\tiny\,\textpm0.08}  \\
    \multirow{2}{*}{LSTM$^{*}$}   & Base    & 0.017{\tiny\,\textpm0.0009} & 0.029{\tiny\,\textpm0.0003} & 0.991{\tiny\,\textpm0.0002} & 3.94{\tiny\,\textpm0.10}  \\
                                  & +TSF    & 0.018{\tiny\,\textpm0.0008} & 0.030{\tiny\,\textpm0.0013} & 0.991{\tiny\,\textpm0.0008} & 4.24{\tiny\,\textpm0.03}  \\
    \multirow{2}{*}{Transformer}  & Base    & 0.020{\tiny\,\textpm0.0011} & 0.033{\tiny\,\textpm0.0012} & 0.990{\tiny\,\textpm0.0008} & 6.49{\tiny\,\textpm0.08}  \\
                                  & +TSF    & 0.020{\tiny\,\textpm0.0007} & 0.033{\tiny\,\textpm0.0007} & 0.989{\tiny\,\textpm0.0004} & 7.76{\tiny\,\textpm0.23}  \\
    \multirow{2}{*}{Informer$^{*}$} & Base    & 0.027{\tiny\,\textpm0.0011} & 0.044{\tiny\,\textpm0.0033} & 0.981{\tiny\,\textpm0.0028} & 58.83{\tiny\,\textpm0.26} \\
                                  & +TSF    & 0.026{\tiny\,\textpm0.0021} & 0.041{\tiny\,\textpm0.0020} & 0.983{\tiny\,\textpm0.0016} & 60.28{\tiny\,\textpm0.75} \\
    \multirow{2}{*}{Mamba$^{*}$}  & Base    & 0.021{\tiny\,\textpm0.0006} & 0.034{\tiny\,\textpm0.0014} & 0.989{\tiny\,\textpm0.0009} & 18.38{\tiny\,\textpm0.23} \\
                                  & +TSF    & 0.018{\tiny\,\textpm0.0009} & 0.031{\tiny\,\textpm0.0008} & 0.990{\tiny\,\textpm0.0005} & 19.66{\tiny\,\textpm0.49} \\
    \multirow{2}{*}{iTransformer$^{*}$} & Base    & 0.020{\tiny\,\textpm0.0020} & 0.033{\tiny\,\textpm0.0026} & 0.989{\tiny\,\textpm0.0020} & 6.81{\tiny\,\textpm0.07}  \\
                                  & +TSF    & 0.023{\tiny\,\textpm0.0014} & 0.039{\tiny\,\textpm0.0036} & 0.985{\tiny\,\textpm0.0030} & 8.19{\tiny\,\textpm0.44}  \\
    \multirow{2}{*}{PatchTST$^{*}$} & Base    & 0.036{\tiny\,\textpm0.0060} & 0.062{\tiny\,\textpm0.0123} & 0.961{\tiny\,\textpm0.0165} & 47.26{\tiny\,\textpm6.44} \\
                                  & +TSF    & 0.035{\tiny\,\textpm0.0023} & 0.059{\tiny\,\textpm0.0035} & 0.967{\tiny\,\textpm0.0040} & 54.26{\tiny\,\textpm9.30} \\
    \multirow{2}{*}{ModernTCN$^{*}$} & Base    & 0.025{\tiny\,\textpm0.0024} & 0.038{\tiny\,\textpm0.0025} & 0.986{\tiny\,\textpm0.0022} & 8.92{\tiny\,\textpm0.32}  \\
                                  & +TSF    & 0.022{\tiny\,\textpm0.0005} & 0.036{\tiny\,\textpm0.0011} & 0.987{\tiny\,\textpm0.0007} & 10.58{\tiny\,\textpm0.12} \\
    \bottomrule
  \end{tabular}
\end{table}

\begin{table}[t]
  \centering
  % 浓密脱水数据集上的定量对比。$^{*}$ 表示 MAE 的 $p<0.05$。
  \caption{Quantitative comparison on the thickener dewatering dataset. $^{*}$ denotes $p<0.05$ for MAE.}
  \label{tab:main_thickener}
  \setlength{\tabcolsep}{2pt}\scriptsize
  \begin{tabular}{llcccr}
    \toprule
    Backbone                      & Setting & MAE$\downarrow$             & RMSE$\downarrow$            & R$^2$ $\uparrow$            & \multicolumn{1}{c}{\makecell{Inference Time\\($\mu$s/sample)$\downarrow$}} \\
    \midrule
    \multirow{2}{*}{GRU}          & Base    & 0.013{\tiny\,\textpm0.0001} & 0.030{\tiny\,\textpm0.0001} & 0.226{\tiny\,\textpm0.0036} & 5.31{\tiny\,\textpm0.11}   \\
                                  & +TSF    & 0.013{\tiny\,\textpm0.0000} & 0.030{\tiny\,\textpm0.0000} & 0.226{\tiny\,\textpm0.0005} & 7.86{\tiny\,\textpm0.14}   \\
    \multirow{2}{*}{LSTM$^{*}$}   & Base    & 0.014{\tiny\,\textpm0.0001} & 0.030{\tiny\,\textpm0.0000} & 0.221{\tiny\,\textpm0.0012} & 5.02{\tiny\,\textpm0.07}   \\
                                  & +TSF    & 0.013{\tiny\,\textpm0.0001} & 0.030{\tiny\,\textpm0.0000} & 0.223{\tiny\,\textpm0.0014} & 7.88{\tiny\,\textpm0.20}   \\
    \multirow{2}{*}{Transformer$^{*}$} & Base    & 0.015{\tiny\,\textpm0.0000} & 0.030{\tiny\,\textpm0.0000} & 0.209{\tiny\,\textpm0.0012} & 14.05{\tiny\,\textpm0.08} \\
                                  & +TSF    & 0.014{\tiny\,\textpm0.0003} & 0.030{\tiny\,\textpm0.0001} & 0.214{\tiny\,\textpm0.0075} & 16.89{\tiny\,\textpm0.22} \\
    \multirow{2}{*}{Informer$^{*}$} & Base    & 0.014{\tiny\,\textpm0.0000} & 0.030{\tiny\,\textpm0.0000} & 0.204{\tiny\,\textpm0.0015} & 133.67{\tiny\,\textpm9.66} \\
                                  & +TSF    & 0.014{\tiny\,\textpm0.0002} & 0.030{\tiny\,\textpm0.0000} & 0.208{\tiny\,\textpm0.0015} & 140.81{\tiny\,\textpm0.50} \\
    \multirow{2}{*}{Mamba$^{*}$} & Base    & 0.015{\tiny\,\textpm0.0001} & 0.030{\tiny\,\textpm0.0000} & 0.196{\tiny\,\textpm0.0017} & 37.13{\tiny\,\textpm0.59} \\
                                  & +TSF    & 0.015{\tiny\,\textpm0.0002} & 0.030{\tiny\,\textpm0.0000} & 0.201{\tiny\,\textpm0.0019} & 40.48{\tiny\,\textpm0.28} \\
    \multirow{2}{*}{iTransformer$^{*}$} & Base    & 0.014{\tiny\,\textpm0.0002} & 0.031{\tiny\,\textpm0.0000} & 0.193{\tiny\,\textpm0.0019} & 14.21{\tiny\,\textpm0.02} \\
                                  & +TSF    & 0.014{\tiny\,\textpm0.0001} & 0.030{\tiny\,\textpm0.0000} & 0.210{\tiny\,\textpm0.0025} & 17.15{\tiny\,\textpm0.18} \\
    \multirow{2}{*}{PatchTST$^{*}$} & Base    & 0.014{\tiny\,\textpm0.0001} & 0.030{\tiny\,\textpm0.0000} & 0.200{\tiny\,\textpm0.0020} & 14.63{\tiny\,\textpm0.09} \\
                                  & +TSF    & 0.015{\tiny\,\textpm0.0001} & 0.031{\tiny\,\textpm0.0001} & 0.182{\tiny\,\textpm0.0037} & 17.66{\tiny\,\textpm0.03} \\
    \multirow{2}{*}{ModernTCN}    & Base    & 0.015{\tiny\,\textpm0.0001} & 0.030{\tiny\,\textpm0.0000} & 0.204{\tiny\,\textpm0.0006} & 14.11{\tiny\,\textpm0.10} \\
                                  & +TSF    & 0.015{\tiny\,\textpm0.0001} & 0.030{\tiny\,\textpm0.0000} & 0.201{\tiny\,\textpm0.0017} & 17.19{\tiny\,\textpm0.06} \\
    \bottomrule
  \end{tabular}
\end{table}

\begin{table}[t]
  \centering
  % IndPenSim 数据集上的定量对比。$^{*}$ 表示 MAE 的 $p<0.05$。
  \caption{Quantitative comparison on the IndPenSim dataset. $^{*}$ denotes $p<0.05$ for MAE.}
  \label{tab:main_indpensim}
  \setlength{\tabcolsep}{2pt}\scriptsize
  \begin{tabular}{llcccr}
    \toprule
    Backbone                      & Setting & MAE$\downarrow$             & RMSE$\downarrow$            & R$^2$ $\uparrow$            & \multicolumn{1}{c}{\makecell{Inference Time\\($\mu$s/sample)$\downarrow$}} \\
    \midrule
    \multirow{2}{*}{GRU$^{*}$}    & Base    & 2.393{\tiny\,\textpm0.0879} & 3.656{\tiny\,\textpm0.0416} & 0.828{\tiny\,\textpm0.0043} & 7.26{\tiny\,\textpm0.18}   \\
                                  & +TSF    & 2.355{\tiny\,\textpm0.1587} & 3.582{\tiny\,\textpm0.2488} & 0.835{\tiny\,\textpm0.0182} & 8.66{\tiny\,\textpm0.22}   \\
    \multirow{2}{*}{LSTM$^{*}$}   & Base    & 2.520{\tiny\,\textpm0.1771} & 4.046{\tiny\,\textpm0.3063} & 0.788{\tiny\,\textpm0.0390} & 7.36{\tiny\,\textpm0.12}   \\
                                  & +TSF    & 2.481{\tiny\,\textpm0.1313} & 3.815{\tiny\,\textpm0.1164} & 0.813{\tiny\,\textpm0.0112} & 9.22{\tiny\,\textpm0.31}   \\
    \multirow{2}{*}{Transformer$^{*}$} & Base    & 2.606{\tiny\,\textpm0.2692} & 3.875{\tiny\,\textpm0.3615} & 0.806{\tiny\,\textpm0.0390} & 14.91{\tiny\,\textpm0.05} \\
                                  & +TSF    & 2.362{\tiny\,\textpm0.2326} & 3.533{\tiny\,\textpm0.2829} & 0.839{\tiny\,\textpm0.0234} & 17.28{\tiny\,\textpm0.59} \\
    \multirow{2}{*}{Informer$^{*}$} & Base    & 2.970{\tiny\,\textpm0.7526} & 4.058{\tiny\,\textpm0.8541} & 0.784{\tiny\,\textpm0.0830} & 135.93{\tiny\,\textpm1.29} \\
                                  & +TSF    & 2.604{\tiny\,\textpm0.4801} & 3.676{\tiny\,\textpm0.4398} & 0.824{\tiny\,\textpm0.0385} & 138.44{\tiny\,\textpm1.76} \\
    \multirow{2}{*}{Mamba$^{*}$}  & Base    & 2.884{\tiny\,\textpm0.1051} & 4.369{\tiny\,\textpm0.1462} & 0.755{\tiny\,\textpm0.0189} & 40.69{\tiny\,\textpm1.09} \\
                                  & +TSF    & 2.859{\tiny\,\textpm0.2697} & 4.301{\tiny\,\textpm0.2041} & 0.762{\tiny\,\textpm0.0248} & 44.57{\tiny\,\textpm0.59} \\
    \multirow{2}{*}{iTransformer$^{*}$} & Base    & 3.876{\tiny\,\textpm0.4684} & 5.498{\tiny\,\textpm0.5037} & 0.609{\tiny\,\textpm0.0749} & 15.79{\tiny\,\textpm0.32} \\
                                  & +TSF    & 2.913{\tiny\,\textpm0.5174} & 4.073{\tiny\,\textpm0.4759} & 0.785{\tiny\,\textpm0.0589} & 19.07{\tiny\,\textpm0.66} \\
    \multirow{2}{*}{PatchTST$^{*}$} & Base    & 2.500{\tiny\,\textpm0.3471} & 3.710{\tiny\,\textpm0.4918} & 0.821{\tiny\,\textpm0.0493} & 213.10{\tiny\,\textpm2.40} \\
                                  & +TSF    & 2.964{\tiny\,\textpm0.4807} & 4.090{\tiny\,\textpm0.3999} & 0.784{\tiny\,\textpm0.0380} & 214.09{\tiny\,\textpm0.62} \\
    \multirow{2}{*}{ModernTCN$^{*}$} & Base    & 2.320{\tiny\,\textpm0.3948} & 3.515{\tiny\,\textpm0.3756} & 0.840{\tiny\,\textpm0.0362} & 15.24{\tiny\,\textpm0.31} \\
                                  & +TSF    & 2.028{\tiny\,\textpm0.0783} & 3.214{\tiny\,\textpm0.0800} & 0.867{\tiny\,\textpm0.0082} & 18.48{\tiny\,\textpm1.03} \\
    \bottomrule
  \end{tabular}
\end{table}

\begin{table}[t]
  \centering
  % Tennessee Eastman Process 数据集上的定量对比。结果由五个固定随机种子汇总得到；$^{*}$ 表示 MAE 的 $p<0.05$。
  \caption{Quantitative comparison on the Tennessee Eastman Process dataset. Results are aggregated over five fixed random seeds; $^{*}$ denotes $p<0.05$ for MAE.}
  \label{tab:main_tep}
  \setlength{\tabcolsep}{2pt}\scriptsize
  \begin{tabular}{llcccr}
    \toprule
    Backbone & Setting & MAE$\downarrow$ & RMSE$\downarrow$ & R$^2$ $\uparrow$ & \multicolumn{1}{c}{\makecell{Inference Time\\($\mu$s/sample)$\downarrow$}} \\
    \midrule
    \multirow{2}{*}{GRU$^{*}$}   & Base & 0.4377{\tiny\,\textpm0.0042} & 0.5614{\tiny\,\textpm0.0069} & 0.0561{\tiny\,\textpm0.0231}  & 5.20{\tiny\,\textpm0.01}   \\
                                  & +TSF & 0.4297{\tiny\,\textpm0.0026} & 0.5466{\tiny\,\textpm0.0052} & 0.1051{\tiny\,\textpm0.0171}  & 7.90{\tiny\,\textpm0.04}   \\
    \multirow{2}{*}{LSTM$^{*}$}  & Base & 0.4377{\tiny\,\textpm0.0043} & 0.5602{\tiny\,\textpm0.0064} & 0.0603{\tiny\,\textpm0.0213}  & 5.35{\tiny\,\textpm0.09}   \\
                                  & +TSF & 0.4319{\tiny\,\textpm0.0027} & 0.5512{\tiny\,\textpm0.0060} & 0.0900{\tiny\,\textpm0.0200}  & 8.12{\tiny\,\textpm0.06}   \\
    \multirow{2}{*}{Transformer$^{*}$} & Base & 0.4423{\tiny\,\textpm0.0065} & 0.5717{\tiny\,\textpm0.0133} & 0.0211{\tiny\,\textpm0.0448}  & 14.75{\tiny\,\textpm0.09}  \\
                                  & +TSF & 0.4320{\tiny\,\textpm0.0014} & 0.5521{\tiny\,\textpm0.0010} & 0.0871{\tiny\,\textpm0.0032}  & 17.07{\tiny\,\textpm0.15}  \\
    \multirow{2}{*}{Informer$^{*}$} & Base & 0.4379{\tiny\,\textpm0.0036} & 0.5653{\tiny\,\textpm0.0082} & 0.0430{\tiny\,\textpm0.0272}  & 141.35{\tiny\,\textpm69.41} \\
                                  & +TSF & 0.4307{\tiny\,\textpm0.0044} & 0.5475{\tiny\,\textpm0.0073} & 0.1023{\tiny\,\textpm0.0244}  & 139.36{\tiny\,\textpm29.46} \\
    \multirow{2}{*}{Mamba$^{*}$} & Base & 0.4569{\tiny\,\textpm0.0017} & 0.5814{\tiny\,\textpm0.0013} & -0.0124{\tiny\,\textpm0.0043} & 40.73{\tiny\,\textpm1.25}  \\
                                  & +TSF & 0.4551{\tiny\,\textpm0.0021} & 0.5784{\tiny\,\textpm0.0046} & -0.0021{\tiny\,\textpm0.0162} & 41.29{\tiny\,\textpm1.27}  \\
    \multirow{2}{*}{iTransformer$^{*}$} & Base & 0.4538{\tiny\,\textpm0.0043} & 0.5929{\tiny\,\textpm0.0085} & -0.0527{\tiny\,\textpm0.0297} & 14.03{\tiny\,\textpm0.26}  \\
                                  & +TSF & 0.4394{\tiny\,\textpm0.0016} & 0.5653{\tiny\,\textpm0.0026} & 0.0431{\tiny\,\textpm0.0087}  & 16.26{\tiny\,\textpm0.20}  \\
    \multirow{2}{*}{PatchTST$^{*}$} & Base & 0.4565{\tiny\,\textpm0.0047} & 0.5988{\tiny\,\textpm0.0107} & -0.0737{\tiny\,\textpm0.0378} & 20.27{\tiny\,\textpm0.08}  \\
                                  & +TSF & 0.4336{\tiny\,\textpm0.0039} & 0.5550{\tiny\,\textpm0.0076} & 0.0775{\tiny\,\textpm0.0257}  & 20.87{\tiny\,\textpm0.01}  \\
    \multirow{2}{*}{ModernTCN$^{*}$} & Base & 0.4434{\tiny\,\textpm0.0053} & 0.5638{\tiny\,\textpm0.0070} & 0.0481{\tiny\,\textpm0.0234}  & 13.18{\tiny\,\textpm0.22}  \\
                                  & +TSF & 0.4299{\tiny\,\textpm0.0064} & 0.5452{\tiny\,\textpm0.0090} & 0.1098{\tiny\,\textpm0.0298}  & 15.71{\tiny\,\textpm0.10}  \\
    \bottomrule
  \end{tabular}
\end{table}

% +TSF 在具有明确工况变化的两个公开过程上表现出最一致的收益。它在 IndPenSim 的八种主干中有七种降低了 MAE，并在 TEP 的八种主干上全部降低了 MAE。这些改进覆盖循环、注意力、状态空间和卷积模型，表明该优势不依赖特定主干。IndPenSim 只使用 recipe-controlled 批次训练，而多数测试批次来自未见过的操作员控制、APC 或故障工况；TEP 划分则同时隔离独立轨迹和工况。在这些划分下，任务感知的变量语义补充了有限数值样本所学的关系，并提高了工况变化时的泛化能力。
The most consistent gains of +TSF appear on the two public processes with explicit operating-condition shifts. It reduces MAE for seven of the eight IndPenSim backbones and all eight TEP backbones. These improvements span recurrent, attention-based, state-space, and convolutional models, showing that the advantage is not tied to a specific backbone. IndPenSim uses only recipe-controlled batches for training, while most test batches follow unseen operator-controlled, APC, or fault regimes; the TEP split isolates both independent trajectories and operating conditions. Under these splits, task-aware variable semantics complement the relations learned from limited numerical samples and improve generalization when operating conditions change.

% 在预测和延迟软测量任务上，+TSF 的平均 MAE 降幅为 3.6\%。对于当前时刻软测量的浓密脱水任务，其收益较小且更依赖主干。全部 32 个数据集--主干组合的宏平均 MAE 降幅为 2.9\%。钢包预热测试集由缺少显式场景标签的完整保留过程组成，浓密脱水测试集则包含 mild-jitter 训练工况之外的远域迁移。当变量级任务语义能够补充有限训练覆盖时，TSF 最有价值。
Across the forecasting and delayed soft-sensing tasks, +TSF reduces MAE by 3.6\% on average. Its gain is smaller and more backbone-dependent for current-step soft sensing on thickener dewatering. The ladle test set consists of complete held-out processes without explicit scenario labels, while thickener dewatering includes far-domain shifts beyond the mild-jitter training regime. The macro-average MAE reduction across all 32 dataset--backbone pairs is 2.9\%. TSF is most valuable when variable-level task semantics complement limited training coverage.

% +TSF 的在线开销保持较小。四个数据集的平均增量分别约为 1.8、3.5、2.4 和 1.5\,$\mu$s/sample，最大增量低于 8\,$\mu$s/sample。
The online overhead of +TSF remains small. The average increments are about 1.8, 3.5, 2.4, and 1.5\,$\mu$s/sample on the four datasets, and the largest increment is below 8\,$\mu$s/sample.

\subsection{Qualitative Analysis}\label{sec:4-4}

\begin{figure*}[t]
  \centering
  \includegraphics[width=181mm]{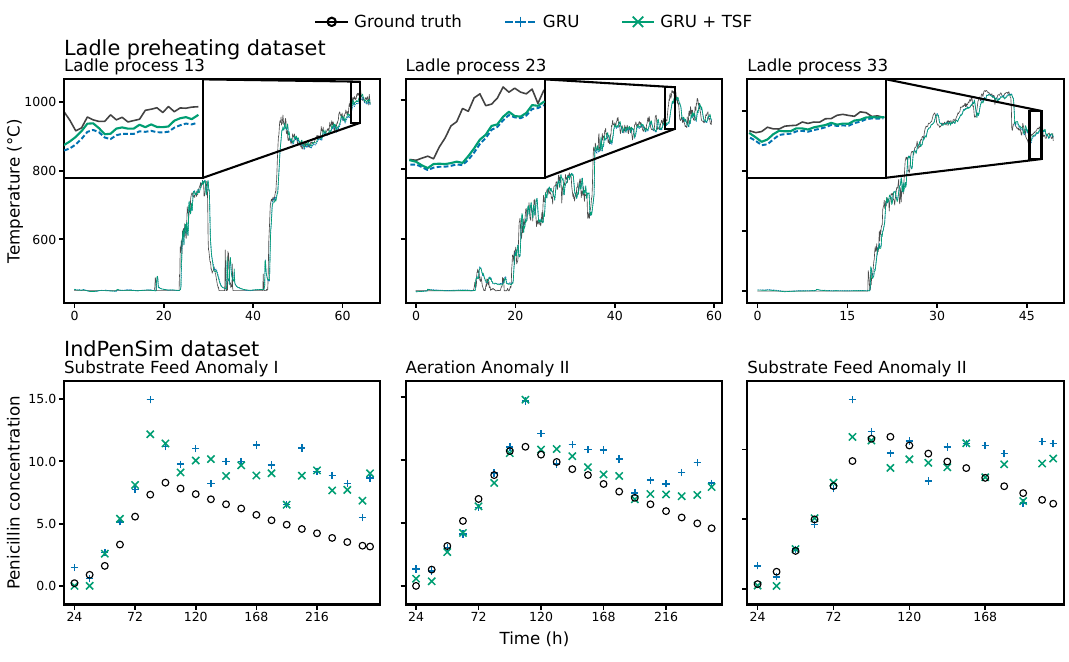}
  % 钢包烘烤和 IndPenSim 上的定性对比。上排给出三个钢包测试过程的第五步钢包温度预测，即 25 分钟后预测；插图放大高温区域。下排给出三个 IndPenSim 异常场景的化验时刻青霉素浓度估计。
  \caption{Qualitative comparison on ladle preheating and IndPenSim. The top row gives fifth-step ladle-temperature forecasts, 25 min ahead, for three ladle processes; insets enlarge high-temperature regions. The bottom row gives assay-time penicillin-concentration estimates for three IndPenSim anomaly cases.}
  \label{fig:qualitative_results}
\end{figure*}

% \cref{fig:qualitative_results} 比较两个代表性任务上的 GRU 与 GRU + TSF。
\cref{fig:qualitative_results} compares GRU and GRU + TSF on two representative tasks.

% 在钢包烘烤上，GRU + TSF 在三条曲线中都更贴近实测温度，尤其减小了加热上升段和高温平台段的持续误差。在 IndPenSim 上，GRU + TSF 在补料和通气异常下使估计值更接近化验真实值，并减小峰值附近和峰值后的偏移。这些可见修正与定量收益一致，也使预测更贴近阶段切换、峰值和平台等关键过程形态。
On ladle preheating, GRU + TSF stays closer to the measured temperature in all three curves, especially by reducing sustained errors during heating rises and high-temperature plateaus. On IndPenSim, GRU + TSF keeps estimates closer to the assay measurements under substrate-feed and aeration anomalies and reduces offsets around peak and post-peak regions. These visible corrections are consistent with the quantitative gains and keep the predictions closer to key process patterns such as stage changes, peaks, and plateaus.

\subsection{Semantic Baselines and Ablation Studies}\label{sec:4-5}

% \cref{tab:semantic_baselines} 比较不同语义来源和集成方式。固定文本基线直接拼接冻结的变量文本嵌入。适配后的 TEST 保留文本原型对齐、软提示和冻结 LLM，并使用相同的 GRU 或 ModernTCN 作为时序编码器；它替代 TSF 输入适配器且不使用 $V$。
\cref{tab:semantic_baselines} compares different semantic sources and integration routes. The fixed-text baseline directly concatenates frozen variable-text embeddings. Adapted TEST retains text-prototype alignment, soft prompts, and a frozen LLM while using the same GRU or ModernTCN as its time-series encoder; it replaces the TSF input adapter and does not use $V$.

\begin{table*}[t]
  \centering
  % 钢包烘烤和 IndPenSim 上的语义基线 MAE 比较。
  \caption{MAE comparison of semantic baselines on ladle preheating and IndPenSim.}
  \label{tab:semantic_baselines}
  {\setlength{\tabcolsep}{2.5pt}\scriptsize
    \begin{tabular}{llcccc}
      \toprule
      \multirow{2}{*}[-0.8ex]{Method} & \multirow{2}{*}[-0.8ex]{Semantic route} & \multicolumn{2}{c}{Ladle Preheating} & \multicolumn{2}{c}{IndPenSim} \\
      \cmidrule(l{2pt}r{2pt}){3-4}\cmidrule(l{2pt}r{0pt}){5-6}
                                      &                                      & GRU                            & ModernTCN                      & GRU                         & ModernTCN                   \\
      \midrule
      Raw backbone                    & None                                 & 0.018{\tiny\,\textpm0.0001} & 0.025{\tiny\,\textpm0.0024} & 2.393{\tiny\,\textpm0.0879} & 2.320{\tiny\,\textpm0.3948} \\
      Fixed-text concatenation        & Frozen names and descriptions       & 0.016{\tiny\,\textpm0.0006} & 0.024{\tiny\,\textpm0.0012} & 2.465{\tiny\,\textpm0.1179} & 2.132{\tiny\,\textpm0.1195} \\
      Metadata + TSF                  & Structured metadata                  & 0.017{\tiny\,\textpm0.0005} & 0.024{\tiny\,\textpm0.0013} & 2.465{\tiny\,\textpm0.1102} & 2.043{\tiny\,\textpm0.0900} \\
      Adapted TEST~\cite{sun2024test} & Text-prototype alignment + frozen LLM & 0.017{\tiny\,\textpm0.0004} & 0.023{\tiny\,\textpm0.0008} & 2.417{\tiny\,\textpm0.1301} & 2.135{\tiny\,\textpm0.1006} \\
      Full TSF                         & Validated LLM semantic cards         & 0.016{\tiny\,\textpm0.0008} & 0.022{\tiny\,\textpm0.0005} & 2.355{\tiny\,\textpm0.1587} & 2.028{\tiny\,\textpm0.0783} \\
      \bottomrule
    \end{tabular}
  }
\end{table*}

\begin{figure}[t]
  \centering
  \includegraphics[width=88mm]{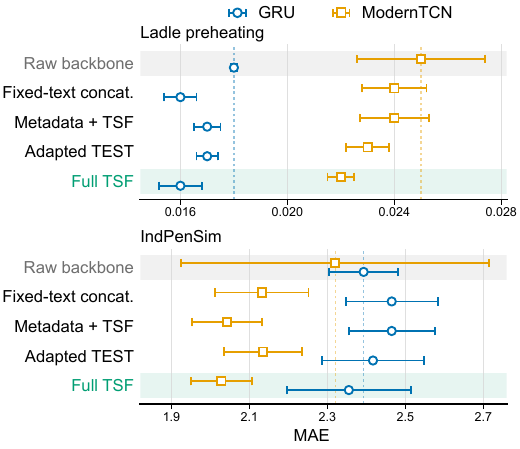}
  % 钢包烘烤和 IndPenSim 上语义基线的 MAE 可视化。标记和水平误差线分别表示均值和标准差。
  \caption{MAE visualization of semantic baselines on ladle preheating and IndPenSim. Markers and horizontal error bars denote the mean and standard deviation, respectively.}
  \label{fig:semantic_baselines_mae}
\end{figure}

% 如 \cref{fig:semantic_baselines_mae} 所示，固定文本拼接和适配后的 TEST 可以改善部分组合，但不能一致复现完整 TSF 的收益。在 IndPenSim--ModernTCN 上，它们的 MAE 分别为 2.132 和 2.135，而 TSF 为 2.028。
As shown in \cref{fig:semantic_baselines_mae}, fixed-text concatenation and adapted TEST can improve some settings, but neither consistently reproduces the gain of full TSF. On IndPenSim--ModernTCN, their MAEs are 2.132 and 2.135, compared with 2.028 for TSF.

% 经过良好整理的结构化元数据在 IndPenSim--ModernTCN 上达到 2.043，接近完整 TSF。这一比较进一步说明了各组件的作用。LLM 利用世界知识和推理，从任务协议及变量文档中构造任务相关语义卡。embedding 模型将验证后的语义卡编码为变量语义方向。TSF 使用当前数值窗口激活这些方向，并以此约束输入映射。常规表示模型编码给定文本，而 LLM 先将分散的任务信息整理为待编码的任务语义。
Well-curated structured metadata reaches 2.043 on IndPenSim--ModernTCN, close to full TSF. This comparison further explains the role of each component. The LLM uses world knowledge and reasoning to construct task-relevant semantic cards from task protocols and variable documents. The embedding model encodes the validated cards into variable-semantic directions. TSF activates these directions with the current numerical window and uses them to constrain the input mapping. A conventional representation model encodes the given text, whereas the LLM first organizes dispersed task information into the task semantics to be encoded.

% \cref{tab:ablation} 进一步隔离自由拟合能力、变量语义对应、原始数值路径和受限因子化。所有变体保持相同的主干、数据划分和训练预算。
\cref{tab:ablation} further isolates free fitting capacity, variable--semantic correspondence, the raw-value path, and constrained factorization. All variants use the same backbones, data splits, and training budgets.

\begin{table*}[t]
  \centering
  % TSF 输入适配层在钢包烘烤和 IndPenSim 上的 MAE 消融实验。
  \caption{MAE ablation study of the TSF input adapter on ladle preheating and IndPenSim.}
  \label{tab:ablation}
  {\setlength{\tabcolsep}{2pt}\scriptsize
    \begin{tabular}{lllcccc}
      \toprule
      \multirow{2}{*}[-0.8ex]{Direction source} & \multirow{2}{*}[-0.8ex]{Direction condition} & \multirow{2}{*}[-0.8ex]{Input mapping} & \multicolumn{2}{c}{Ladle Preheating} & \multicolumn{2}{c}{IndPenSim} \\
      \cmidrule(l{2pt}r{2pt}){4-5}\cmidrule(l{2pt}r{0pt}){6-7}
                                                 &                            &                               & GRU                            & ModernTCN                      & GRU                         & ModernTCN                   \\
      \midrule
      None                                       & --                         & Raw input                     & 0.018{\tiny\,\textpm0.0001} & 0.025{\tiny\,\textpm0.0024} & 2.393{\tiny\,\textpm0.0879} & 2.320{\tiny\,\textpm0.3948} \\
      Random directions                          & Fixed by index             & TSF factorization             & 0.017{\tiny\,\textpm0.0005} & 0.025{\tiny\,\textpm0.0012} & 2.515{\tiny\,\textpm0.0744} & 2.340{\tiny\,\textpm0.2387} \\
      Learnable directions                       & Learned by index           & TSF factorization             & 0.017{\tiny\,\textpm0.0005} & 0.023{\tiny\,\textpm0.0008} & 2.375{\tiny\,\textpm0.1153} & 2.288{\tiny\,\textpm0.1398} \\
      Semantic cards                             & Shuffled                   & TSF factorization             & 0.017{\tiny\,\textpm0.0004} & 0.025{\tiny\,\textpm0.0019} & 2.448{\tiny\,\textpm0.0642} & 2.370{\tiny\,\textpm0.2841} \\
      Semantic cards                             & Conflicted                 & TSF factorization             & 0.017{\tiny\,\textpm0.0005} & 0.024{\tiny\,\textpm0.0010} & 2.430{\tiny\,\textpm0.0963} & 2.206{\tiny\,\textpm0.1233} \\
      Semantic cards                             & Validated                  & Semantic path only            & 0.017{\tiny\,\textpm0.0004} & 0.023{\tiny\,\textpm0.0024} & 2.599{\tiny\,\textpm0.1368} & 2.446{\tiny\,\textpm0.1375} \\
      Semantic cards                             & Validated                  & Full numeric matrix           & 0.017{\tiny\,\textpm0.0004} & 0.023{\tiny\,\textpm0.0005} & 2.389{\tiny\,\textpm0.0233} & 2.429{\tiny\,\textpm0.2069} \\
      Semantic cards                             & Validated                  & TSF factorization             & 0.016{\tiny\,\textpm0.0008} & 0.022{\tiny\,\textpm0.0005} & 2.355{\tiny\,\textpm0.1587} & 2.028{\tiny\,\textpm0.0783} \\
      \bottomrule
    \end{tabular}
  }
\end{table*}

% 随机方向和可学习方向保留了输入适配结构，但在 IndPenSim--ModernTCN 上分别得到 2.340 和 2.288，均不及完整 TSF 的 2.028。这一差距表明，适配结构或自由学习方向本身不足以解释主要收益。
Random and learnable directions retain the input-adaptation structure, but reach 2.340 and 2.288 on IndPenSim--ModernTCN, respectively, compared with 2.028 for full TSF. This gap indicates that the adapter structure or freely learned directions alone are insufficient to explain the main gain.

% 语义质量和正确对应决定了这种收益是否可靠。在 IndPenSim--ModernTCN 上，引入受控冲突后 MAE 从 2.028 升至 2.206，打乱变量对应后进一步升至 2.370。这一结果也支持在构建 $V$ 前验证语义卡。
Semantic quality and correct correspondence determine whether this gain remains reliable. On IndPenSim--ModernTCN, MAE rises from 2.028 to 2.206 under controlled conflicts and to 2.370 after shuffling variable correspondence. This result also supports validating semantic cards before constructing $V$.

% 原始数值路径和受限因子化保持了清晰的输入结构。去掉原始数值路径会削弱 IndPenSim 上的两个主干，自由完整矩阵在 IndPenSim--ModernTCN 上的 MAE 为 2.429。钢包烘烤上的差异较小，说明这些设计的作用会随任务和主干变化。语义基线和消融结果共同表明，TSF 将经过验证的语义方向与受限输入映射结合。语义方向引导跨变量混合，原始数值路径则保留每个变量的数值信息。
The raw-value path and constrained factorization preserve a clear input structure. Removing the raw-value path weakens both IndPenSim backbones, while the free full matrix gives an MAE of 2.429 on IndPenSim--ModernTCN. Differences on ladle preheating are smaller, indicating that these design effects remain task- and backbone-dependent. The semantic baselines and ablations together show that TSF combines validated semantic directions with a constrained input mapping. The semantic directions guide cross-variable mixing, while the raw-value path preserves the numerical information of each variable.

\subsection{Task-Semantic Field Visualization}\label{sec:4-6}

\begin{figure*}[t]
  \centering
  \includegraphics[width=181mm]{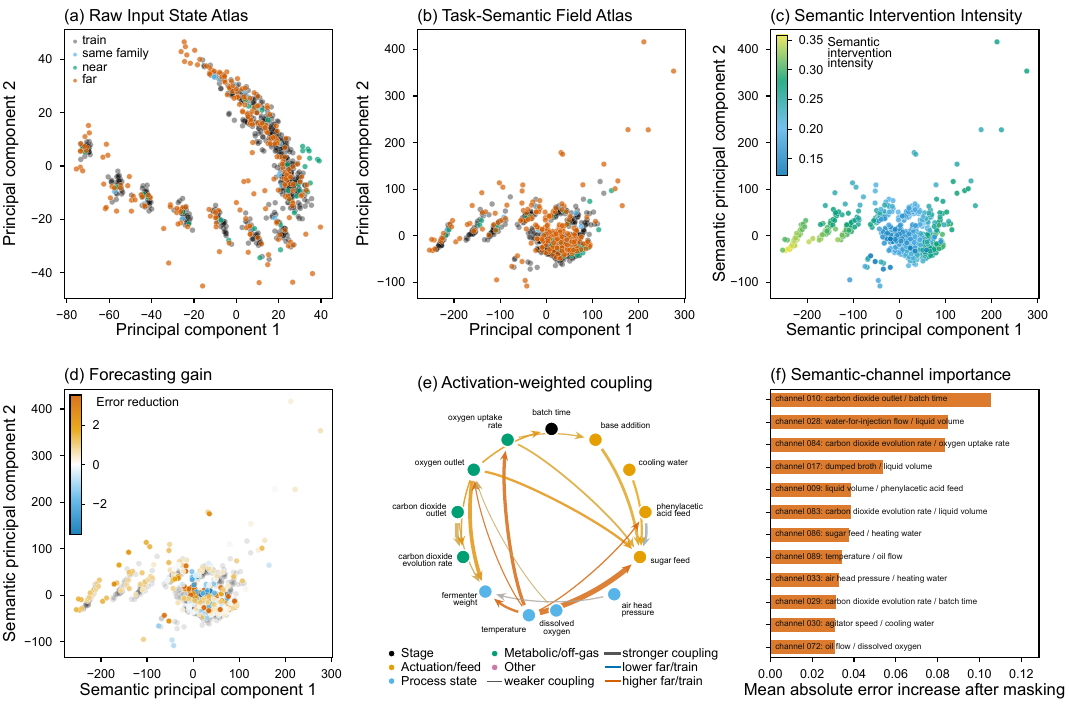}
  % TSF 在 IndPenSim-ModernTCN 上的运行时任务语义场分析。(a)~原始归一化输入窗口的 PCA 状态图。(b)~由 $S=\tilde{X}V$ 形成的运行时任务语义场状态图。(c)~语义路径介入强度 $\rho=\|SB\|_{F}/(\|\tilde{X}D\|_{F}+\|SB\|_{F})$，衡量样本层面两条路径中语义投影的相对 Frobenius 范数幅度。(d)~TSF 相对原始输入主干的样本级绝对误差降低，正值表示 TSF 误差更小。(e)~far-domain 样本中按实际输入激活加权的变量耦合边；节点颜色表示变量角色，边宽表示耦合强度，边颜色表示 far/train 富集方向。(f)~在 far-domain 样本上逐一屏蔽语义通道得到的反事实通道重要性。
  \caption{Runtime task-semantic field analysis on IndPenSim-ModernTCN. (a)~PCA state atlas of raw normalized input windows. (b)~Runtime task-semantic field atlas formed by $S=\tilde{X}V$. (c)~Semantic intervention intensity $\rho=\|SB\|_{F}/(\|\tilde{X}D\|_{F}+\|SB\|_{F})$, measuring the relative Frobenius-norm magnitude of the semantic projection across the two paths at the sample level. (d)~Sample-level absolute-error reduction of TSF over the raw-input backbone, where positive values indicate lower TSF error. (e)~Activation-weighted variable couplings in far-domain samples; node colors denote variable roles, edge widths denote coupling strength, and edge colors denote far/train enrichment. (f)~Counterfactual semantic-channel importance from masking one semantic channel at a time on far-domain samples.}
  \label{fig:semantic_field_visualization}
\end{figure*}

% \cref{fig:semantic_field_visualization} 可视化了 IndPenSim-ModernTCN 运行中的任务语义场。
\cref{fig:semantic_field_visualization} visualizes the task-semantic field in the IndPenSim-ModernTCN run.

% 与原始输入空间相比，任务语义场解释了更高比例的二维主成分方差，并把远域样本展开到更宽的语义区域。语义路径介入强度在该空间中呈现区域差异，说明 $SB$ 的贡献随当前窗口状态变化。对应的误差降低图显示收益分布在多个 far-domain 区域。这表明 TSF 形成的运行时语义结构与实际预测改进一致，而不是简单复制原始数值几何。
Compared with the raw input space, the task-semantic field explains a larger fraction of the two-dimensional principal-component variance and spreads far-domain samples over a broader semantic region. The semantic-path intervention intensity varies across this space, showing that the contribution of $SB$ changes with the current window state. The corresponding error-reduction map shows gains across several far-domain regions. This pattern indicates that the runtime semantic structure formed by TSF is aligned with actual forecasting improvement rather than simply copying the raw numerical geometry.

% 这种结构也可以追踪到过程变量和语义通道。far-domain 激活加权耦合集中在温度、压力、补糖、PAA、加热/冷却、溶氧和尾气变量之间，对应发酵中的热状态、底物供给、前体添加、气液传质和代谢强度。反事实通道置零进一步显示，重要通道与 \texttt{co2\_out}/batch time、water-for-injection/volume、CER/OUR、dumped broth/volume 和 temperature/oil flow 等方向相关。因此，TSF 的输入适配不只是增加容量，而是把训练前固定的变量语义转化为运行时可激活、可干预且与预测收益相关的输入结构。
This structure can also be traced to process variables and semantic channels. The far-domain activation-weighted couplings concentrate around temperature, pressure, sugar feed, PAA feed, heating/cooling, dissolved oxygen, and off-gas variables, which correspond to thermal state, substrate supply, precursor addition, gas-liquid transfer, and metabolic intensity in fermentation. Counterfactual channel masking further shows that important channels are associated with directions such as \texttt{co2\_out}/batch time, water-for-injection/volume, CER/OUR, dumped broth/volume, and temperature/oil flow. TSF therefore does more than add capacity to the input layer. It turns variable semantics fixed before training into a runtime input structure that is activated by the current window, supports intervention, and is associated with forecasting gain.

\subsection{Sensitivity Analysis}\label{sec:4-7}

\begin{figure*}[t]
  \centering
  \includegraphics[width=181mm]{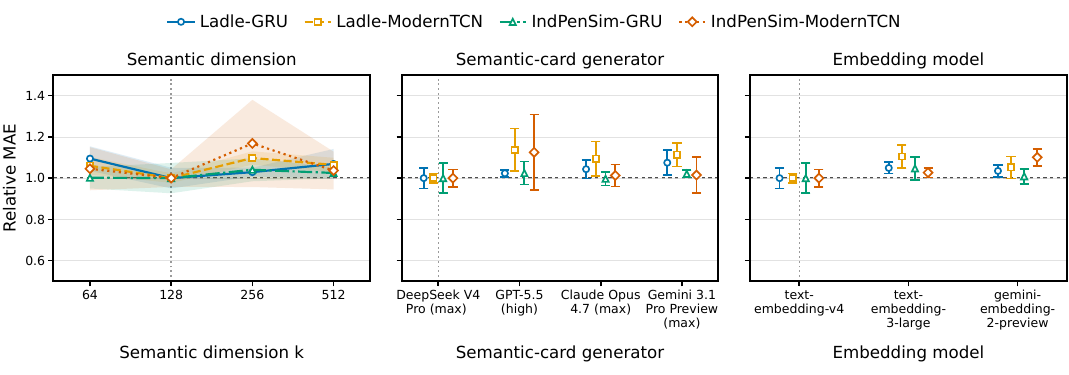}
  % TSF 对语义维度、语义卡生成器和 embedding 模型的敏感性分析。数值表示相对于每个面板默认设置的相对 MAE。阴影带和误差条表示五次运行的标准差；灰色虚线标记默认设置。
  \caption{Sensitivity analysis of TSF across semantic dimension, semantic-card generator, and embedding model. Values are relative MAE with respect to the default setting in each panel. Shaded bands and error bars denote standard deviations over five runs; gray dashed lines mark the default setting.}
  \label{fig:sensitivity_analysis}
\end{figure*}

% 本小节从语义维度、语义卡生成器和 embedding 模型三个因素分析 TSF 的敏感性，结果如 \cref{fig:sensitivity_analysis} 所示。
We analyze TSF sensitivity along three factors: semantic dimension, semantic-card generator, and embedding model, as shown in \cref{fig:sensitivity_analysis}.

% 语义维度并不呈现单调趋势。默认的 $k=128$ 提供了稳定工作点，而逐项最优值会随数据集和主干变化。在 IndPenSim-ModernTCN 上，$k=256$ 的相对 MAE 升至约 1.17，说明过大的语义空间可能降低部分数据集--主干组合的性能。其他组合在不同 $k$ 下的变化更小。因此，我们采用 $k=128$ 作为默认值，以兼顾语义容量和训练稳定性。
The semantic dimension does not show a monotonic trend. The default $k=128$ gives a stable operating point, while the best value changes with the dataset and backbone. On IndPenSim-ModernTCN, the relative MAE at $k=256$ rises to about 1.17, showing that an overly large semantic space can degrade performance for some dataset--backbone pairs. The other pairs are less sensitive to $k$. We therefore use $k=128$ as the default to balance semantic capacity and training stability.

% 固定 text-embedding-v4 后，不同 LLM 生成的语义卡整体都可以支持 TSF。钢包烘烤上的变化较小，IndPenSim-ModernTCN 对语义卡来源更敏感。GPT-5.5 生成的语义卡在该组合上的相对 MAE 高于默认结果，而 Claude Opus 4.7 和 Gemini 3.1 Pro 更接近默认结果。该结果说明 TSF 不绑定单一语义生成器，但分布迁移更大的任务仍会受到语义卡质量影响。
With text-embedding-v4 fixed, semantic cards generated by different LLMs can all support TSF. The changes on ladle preheating are small, while IndPenSim-ModernTCN is more sensitive to the semantic-card source. GPT-5.5 gives a higher relative MAE than the default result for this pair, whereas Claude Opus 4.7 and Gemini 3.1 Pro are closer to the default result. This result indicates that TSF is not tied to a single semantic generator, but tasks with larger distribution shifts remain sensitive to semantic-card quality.

% 替换 embedding 模型会直接改变语义方向空间。两个替代 embedding 在钢包烘烤上保持可比结果，但在 IndPenSim-ModernTCN 上都高于默认设置，且 gemini-embedding-2-preview 的增幅更明显。相比 LLM 替换，embedding 替换更直接地改变变量方向之间的几何关系。
Changing the embedding model directly changes the semantic-direction space. The two alternative embeddings remain competitive on ladle preheating, but both give higher relative MAE than the default on IndPenSim-ModernTCN, with a larger increase for gemini-embedding-2-preview. Compared with LLM replacement, embedding replacement more directly changes the geometry among variable directions.

\subsection{Computational Efficiency}\label{sec:4-8}

% 为评估在线部署成本，\cref{tab:efficiency} 报告在相同主干上加入 TSF 输入适配器后的模型规模、单样本推理时间和峰值内存变化。
To evaluate online deployment cost, \cref{tab:efficiency} reports model size, per-sample inference time, and peak-memory change after adding the TSF input adapter to the same backbone.

\begin{table}[t]
  \centering
  % TSF 输入适配器的在线部署成本。LLM 语义卡和 embedding 生成不计入在线推理。
  \caption{Online deployment cost of the TSF input adapter. LLM semantic-card and embedding generation are excluded from online inference.}
  \label{tab:efficiency}
  \setlength{\tabcolsep}{2pt}\scriptsize
  \begin{tabular}{lrrrrr}
    \toprule
              & \multicolumn{2}{c}{Model Size $\downarrow$} & \multicolumn{3}{c}{Deployment Cost $\downarrow$}                           \\
    \cmidrule(lr){2-3}\cmidrule(lr){4-6}
    Backbone  & \makecell{Backbone\\(\,\texttimes\,10\textsuperscript{3})} & \makecell{Adapter\\(\,\texttimes\,10\textsuperscript{3})} & \makecell{Base Inference\\($\mu$s/sample)} & \makecell{+TSF Inference\\($\mu$s/sample)} & \makecell{Peak RSS\\Increase (MB)} \\
    \midrule
    \multicolumn{6}{l}{\textit{Ladle Preheating}}                                                                                        \\
    GRU       & 88.613                                      & 1.820                                            & 3.52{\tiny\,\textpm0.03}  & 3.73{\tiny\,\textpm0.08}  & 0.22  \\
    ModernTCN & 78.341                                      & 1.820                                            & 8.92{\tiny\,\textpm0.32}  & 10.58{\tiny\,\textpm0.12} & 4.61  \\
    \addlinespace[1pt]
    \multicolumn{6}{l}{\textit{IndPenSim}}                                                                                               \\
    GRU       & 90.817                                      & 2.990                                            & 7.26{\tiny\,\textpm0.18}  & 8.66{\tiny\,\textpm0.22}  & 15.23 \\
    ModernTCN & 78.817                                      & 2.990                                            & 15.24{\tiny\,\textpm0.31} & 18.48{\tiny\,\textpm1.03} & 0.74  \\
    \bottomrule
  \end{tabular}
\end{table}

% 结果表明，TSF 的在线成本较低，因为语义构造被前移到训练前，而在线阶段只保留一个小型输入适配器。该设计在不改变主干推理流程的情况下加入任务语义。在表中所列配置中，+TSF 的最大时间为 18.48\,$\mu$s/sample，远低于工业任务的采样间隔。Peak RSS 变化没有形成新的部署瓶颈。
The results show that TSF has low online cost because semantic construction is moved before training and only a small input adapter remains online. This design adds task semantics without changing the backbone inference pipeline. Among the configurations in the table, the largest +TSF time is 18.48\,$\mu$s/sample, far below the sampling intervals of the industrial tasks. Peak RSS changes do not create a new deployment bottleneck.

\FloatBarrier

\section{Conclusion}\label{sec:conclusion}

% 本文提出 TSF，一个使数值时序模型从过程文档中获得任务感知变量语义的框架。它把 LLM 提炼出的过程知识表示为输入变量与预测目标之间的语义逻辑关系，并由当前数值窗口激活这些关系。因此，变量语义从模型外部文档进入预测过程，并支持模型面对不同预测目标和工况变化时进行适应。
This article presented TSF, a framework that equips numerical time-series models with task-aware variable semantics from process documents. It represents LLM-derived process knowledge as semantic-logical relations between input variables and the prediction target and activates these relations with the current numerical window. Variable semantics therefore move from external documentation into the prediction process and support adaptation to different prediction targets and operating shifts.
% 在多个复杂工业预测和延迟软测量任务上，实验结果表明当变量语义能够补充有限训练覆盖时，TSF 可以降低预测误差。+TSF 的平均 MAE 降幅为 3.6\%；在全部 32 个数据集--主干组合上，宏平均降幅为 2.9\%，最大降幅为 24.9\%。在线适配器只增加约 0.7--4.3k 个参数，额外在线推理开销低于 8\,$\mu$s/sample。这些结果说明，TSF 可以把已有过程文档转化为跨主干和语义生成器的可部署预测收益。
Across multiple complex industrial forecasting and delayed soft-sensing tasks, experimental results show that TSF reduces forecasting error when variable semantics can complement limited training coverage. +TSF reduces MAE by 3.6\% on average. Across all 32 dataset--backbone pairs, the macro-average reduction is 2.9\%, with a maximum reduction of 24.9\%. The online adapter adds only about 0.7--4.3k parameters, with less than 8\,$\mu$s/sample of additional online inference overhead. These results show that TSF can turn existing process documents into deployable forecasting gains across backbones and semantic generators.

% TSF 仍依赖可用的任务协议、变量表和工艺说明。冻结前验证能够处理可识别的字段缺失和记录冲突，但含糊情况仍需要人工确认。未来工作将减少这部分人工工作，并在更长周期的工业部署中评估 TSF。
TSF still depends on usable task protocols, variable tables, and process descriptions. Pre-freezing validation handles identifiable missing fields and record conflicts, but ambiguous cases still require manual confirmation. Future work will reduce this manual effort and evaluate TSF in longer-term industrial deployments.

% 参考文献
\bibliographystyle{IEEEtran}
\bibliography{references}

\end{document}